\documentclass[runningheads,orivec]{llncs}
\usepackage{times}
\usepackage{epsfig}
\usepackage{graphicx}
\usepackage{comment}

\usepackage{amsmath,amssymb}
\usepackage[utf8]{inputenc} 
\usepackage[T1]{fontenc}    
\usepackage{url}   
\usepackage{color}
\usepackage{booktabs}       
\usepackage{amsfonts}       
\usepackage{nicefrac}       
\usepackage{microtype}      

\usepackage{amstext}
\usepackage{tabulary}

\usepackage{amsthm}
\usepackage{amssymb}
\usepackage{bm}
\usepackage{wrapfig}
\usepackage{subfigure}
\usepackage{graphicx}
\usepackage{bbm, comment}
\usepackage{color}
\usepackage{float}
\usepackage{wrapfig}
\usepackage{enumitem}
\usepackage{url}
\usepackage{MnSymbol,wasysym,marvosym,ifsym}
\usepackage{algorithm}
\usepackage{algorithmic}
\usepackage{wrapfig}

\usepackage{mathtools}
\def\setstretch#1{\renewcommand{\baselinestretch}{#1}}
\newcommand{\E}{\mathbb{E}}

\usepackage[pagebackref=true,breaklinks=true,letterpaper=true,colorlinks,bookmarks=false]{hyperref}

\newtheorem{assumption}[theorem]{Assumption}

\newcommand{\tc}{\mathrm{TC}}
\newcommand{\ptc}{\mathrm{PTC}}
\newcommand{\ctc}{\mathrm{CTC}}
\newcommand{\h}{\mathrm{H}}
\newcommand{\kl}{\mathrm{KL}}
\DeclareRobustCommand\onedot{\futurelet\@let@toke\@onedot}
\def\@onedot{\ifx\@let@token.\else\null\fi\xspace}

\def\ie{\emph{i.e}\onedot}

\def\etal{\emph{et al}\onedot}

\def\ECCVSubNumber{6209} 

\begin{document}
\pagestyle{headings}
\mainmatter

\title{TCGM: An Information-Theoretic Framework for Semi-Supervised Multi-Modality Learning}

\titlerunning{TCGM}
%
\author{Xinwei Sun\thanks{Equal Contribution.}\inst{1} \and
Yilun Xu\protect\footnotemark[1] \inst{2,3} \and\\
Peng Cao\inst{2,3} \and Yuqing Kong \inst{2}  \and Lingjing Hu\inst{4}\and  Shanghang Zhang  \inst{5} \and  Yizhou Wang\inst{2,3,6,7}}
\authorrunning{Sun et al.}
%
\institute{Microsoft Research-Asia\\ \email{xinsun@microsoft.com} \and
Center on Frontiers of Computing Studies, Peking University
\and
Dept. of  Computer Science, Peking University
\\ \email{\{xuyilun, caopeng2016, yuqing.kong, yizhou.wang\}@pku.edu.cn} \\
\and 
Yanjing Medical College, Capital Medical University\\ \email{hulj@ccmu.edu.cn} \\
\and UC Berkeley\\ \email{shz@eecs.berkeley.edu} \\\and{Adv. Inst. of Info. Tech, Peking University} \and Deepwise AI Lab }

\definecolor{pink}{rgb}{0.858, 0.188, 0.478}
\newcommand\yk[1]{\textcolor{cyan}{[Yuqing: #1]}}
\newcommand\yw[1]{\textcolor{green}{[Yizhou: #1]}}
\newcommand\yx[1]{\textcolor{red}{[Yilun: #1]}}
\newcommand\xs[1]{\textcolor{blue}{[Xinwei: #1]}}
\newcommand\pc[1]{\textcolor{pink}{[Peng: #1]}}

\maketitle

\begin{abstract}
Fusing data from multiple modalities provides more information to train machine learning systems. However, it is prohibitively expensive and time-consuming to label each modality with a large amount of data, which leads to a crucial problem of semi-supervised multi-modal learning. Existing methods suffer from either ineffective fusion across modalities or lack of theoretical guarantees under proper assumptions. In this paper, we propose a novel information-theoretic approach \-- namely, \textbf{T}otal \textbf{C}orrelation \textbf{G}ain \textbf{M}aximization (TCGM) \--- for semi-supervised multi-modal learning, which is endowed with promising properties: (i) it can utilize effectively the information across different modalities of unlabeled data points to facilitate training classifiers of each modality (ii) it has theoretical guarantee to identify Bayesian classifiers, i.e., the ground truth posteriors of all modalities. Specifically, by maximizing TC-induced loss (namely TC gain) over classifiers of all modalities, these classifiers can cooperatively discover the equivalent class of ground-truth classifiers; and identify the unique ones by leveraging limited percentage of labeled data. We apply our method to various tasks and achieve state-of-the-art results, including the news classification (Newsgroup dataset), emotion recognition (IEMOCAP and MOSI datasets), and disease prediction (Alzheimer’s Disease Neuroimaging Initiative dataset).\newline




\end{abstract}

\section{Introduction}
\begin{figure}[h!]
    \centering
    \includegraphics[width=3.2in]{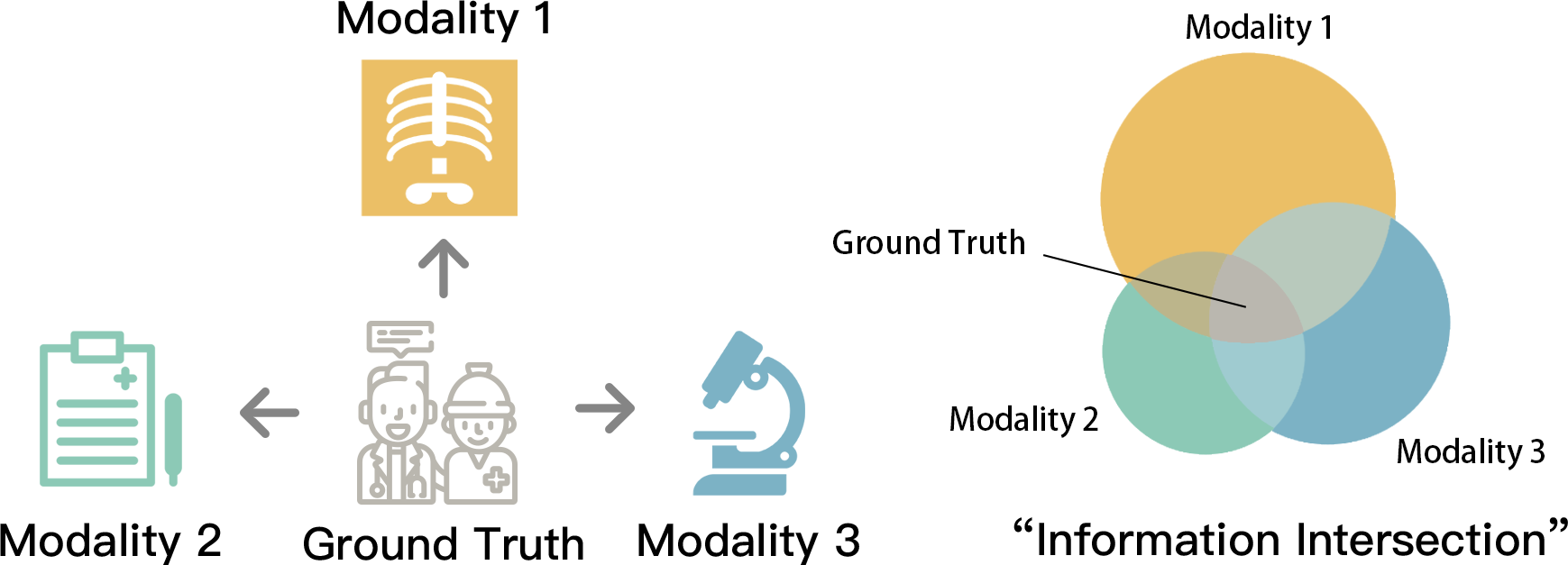}
    \caption{Multiple modalities are independent conditioning on the ground truth; Ground truth is the ``information intersection" of all of the modalities.}
    \label{fig:my_label}
\end{figure}

Learning with data from multiple modalities has the advantage to facilitate information fusion from different perspectives and induce more robust models, compared with only using a single modality. For example, as shown in Figure \ref{fig:my_label}, to diagnose whether a patient has a certain disease or not, we can consult to its X-ray images, look into its medical records, or get results from clinical pathology. However, in many real applications, especially in some difficult ones (e.g. medical diagnosis), annotating such large-scale training data is prohibitively expensive and time-consuming. As a consequence, each modality of data may only contain a small proportion of labeled data from professional annotators, leaving a large proportion of unlabeled. This leads to an essential and challenging problem of semi-supervised multi-modality learning: \emph{how to effectively train accurate classifiers by aggregating unlabeled data of all modalities? }



To achieve this goal, many methods have been proposed in the literature, which can be roughly categorized into two branches: (i) co-training strategy \cite{blum1998combining}; and (ii) learning joint representation across modalities in an unsupervised way \cite{ngiam2011multimodal,sohn2014improved}. These methods suffer from either too strong assumptions or loss of information during fusing. Specifically, the co-training strategy relies largely on the ``compatible'' assumption that the conditional distributions of the data point labels in each modality are the same, which may not be satisfied in the real settings, as self-claimed in \cite{blum1998combining}; while the latter branch of methods fails to capture the higher-order dependency among modalities, hence may end up in learning a trivial solution that maps all the data points to the same representation.

A common belief in multi-modality learning \cite{lewis1998naive,blum1998combining,dasgupta2002pac,leskes2005value} is that conditioning on ground truth label $Y$, these modalities are conditionally independent, as illustrated in Figure \ref{fig:my_label}. For example, to diagnose if one suffers from a certain disease, an efficient way is to leverage as many as modalities that are related to the disease, e.g., X-ray image, medical records and the clinical pathology. Since each modality captures the characteristics of the disease from different aspects, the information extracted from these modalities, in addition to the label, are not necessarily correlated with each other. This suggests that the ground truth label can be regarded as the ``information intersection'' across all the modalities, i.e., the amount of agreement shared by all the modalities.

Inspired by such an assumption and the fact that the \emph{Total Correlation} \cite{studeny1998multiinformation} can measure the amount of information shared by $M$ ($M \geq 2$) variables, in this paper, we propose \textbf{T}otal \textbf{C}orrelation \textbf{G}ain (TCG), which is a function of classifiers of all the modalities, as a surrogate goal for maximization of mutual information, in order to infer the ground-truth labels (i.e., information intersection among these modalities). 
Based on the proposed TCG, we devise an information-theoretic framework called \emph{\textbf{Total Correlation Gain Maximization}} (TCGM) for semi-supervised multi-modal learning.
By maximizing TCG among all the modalities, the classifiers for different modalities cooperatively discover the information intersection across all the modalities.
It can be proved that the optimal classifiers for such a Total Correlation Gain are equivalent to the Bayesian posterior classifiers given each modality under some permutation function. With further leverage of labeled data, we can identify the Bayesian posterior classifiers. Furthermore, we devise an aggregator that employs all the modalities to forecast the labels of data. A simulated experiment is conducted to verify this theoretical result.

We apply TCGM on various tasks: (i) News classification with three pre-processing steps as different modalities, (ii) Emotion recognition with videos, audios, and texts as three modalities and (iii) disease prediction on medical imaging with the Structural magnetic resonance imaging (sMRI) and Positron emission tomography (PET) modalities. On these tasks, our method consistently outperforms the baseline methods especially when a limited percentage of labeled data are provided. To validate the benefit of jointly learning, we visualize that some cases of Alzheimer's Disease whose label are difficult to be predicted via supervised learning with single modality; while our jointly learned single modal classifier is able to correctly classify such hard samples. 





The contributions can be summarized as follows: (i)  We propose a novel information-theoretic approach TCGM for semi-supervised multi-modality learning, which can effectively utilize information across all modalities. By maximizing the total correlation gain among all the modalities, the classifiers for different modalities cooperatively discover the information intersection across all the modalities - the ground truth. (ii)  To the best of our knowledge, \textbf{TCGM} is the first in the literature that can be theoretically proved that, under the conditional independence assumption, it can identify the ground-truth Bayesian classifier given each modality. Further, by aggregating these classifiers, our method can learn the Bayesian classifier given all modalities. (iii) We achieve the state-of-the-art results on various semi-supervised multi-modality tasks including news classification, emotion recognition and disease prediction of medical imaging.

\section{Related Work}
\paragraph{Semi-supervised multi-modal learning} It is commonly believed in the literature that information of label is shared across all modalities. Existing work, which can be roughly categorized into two branches, suffers from either stronger but not reasonable assumptions or failure to capture the information (i.e., label) shared by all modalities. The first branch applies the co-training algorithm proposed by Blum et. al \cite{blum1998combining}. \cite{jones2005learning,balcan2005co,christoudias2006co,levin2003unsupervised,cheng2015semi} use weak classifiers trained by the labeled data from each modality to bootstrap each other by generating labels for the unlabeled data. However, the underlying compatible condition of such a method, which assumes the same conditional distributions for data point labels in each modality, may not be consistent with the real settings.

The second branch of work \cite{ngiam2011multimodal,sohn2014improved,chandar2016correlational,wang2018efficient,chang2018scalable,kim2017multi} centers on learning joint representations that project unimodal representations all together into a multi-modal space in an unsupervised way and then using the labeled data from each modality to train a classifier to predict the label of the learned joint representation. A representative of such a framework is the soft-Hirschfeld-Gebelein-R\'enyi (HGR) framework \cite{wang2018efficient}, which proposed to maximize the correlation among non-linear representations of each modality. However, HGR only measures the linear dependence between pair modalities, since it follows the principle of maximizing the correlation between features of different modalities. In contrast, our framework, i.e., Total Correlation Gain Maximization can pursue information about higher-order dependence. Due to the above reasons, both branches can not avoid learning a naive solution that classifies all data points into the same class.


To overcome these limitations, we propose an information-theoretic loss function based on Total Correlation which can not only require the assumption in the first branch of work but also can be able to identify the ground-truth label, which is the information intersection among these modalities. Therefore, our method can avoid the trivial solution and can learn the optimal, i.e., the Bayesian Posterior classifiers of each modality. 




\paragraph{Total Correlation/Mutual information maximization} Total Correlation \cite{studeny1998multiinformation}, as an extension of Mutual Information, measures the amount of information shared by $M$ $(M\geq 2)$ variables. There are several works in the literature that have combined Mutual Information $(M=2)$ with deep learning algorithms and have shown superior performance on various tasks. Belghazi \etal \cite{belghazi2018mine} presents a mutual information neural estimator, which are utilized in a handful of applications based on the mutual information maximization (e.g., unsupervised learning of representations \cite{hjelm2018learning}, learning node representations within graph-structured data \cite{velivckovic2018deep}). Kong and Schoenebeck \cite{kong2018water} provide another mutual information estimator in the co-training framework for the peer prediction mechanism, which has been combined with deep neural networks for crowdsourcing \cite{cao2018max}. Xu et.al \cite{Xu2020ATO} proposes an alternative definition of information, which is more effective for structure learning. However, those three estimators can only be applied to two-view settings. To the best of our knowledge, there are no similar studies that focus on a general number of modalities, which is very often in real applications. In this paper, we propose to leverage Total Correlation to fill in such a gap. 


\section{Preliminaries}
\paragraph{Notations} Given a random variable $X$, $\mathcal{X}$ denotes its realization space and $x \in \mathcal{X}$ denotes an instance. The $\mbox{P}_{\mathcal{X}}$ denotes the probability distribution function over $\mathcal{X}$ and $p(x) := d\mbox{P}_{\mathcal{X}}(x)$ denotes the density function w.r.t the Lebesgue measure. Further, given a finite set $\mathcal{X}$, $\Delta_{\mathcal{X}} $ denotes the set of all distributions over $\mathcal{X}$. For every integer $M$, $[M]$ denotes the set $\{1,2,\dots,M\}$. For a vector $\mathbf{v}$, $v_i$ denotes its $i$-th element. 

\paragraph{Total Correlation} The Total Correlation (TC), as an extension of mutual information, measures the ``amount of information" shared by $M$ ($\geq 2$) random variables: 
\begin{equation}
    \label{eq:TC}
    \tc(X^1,...,X^m) = \sum_{i=1}^M \h(X^i) - \h(X^1,...,X^M),
\end{equation}
where $\h$ is the Shannon entropy. As defined, the $\tc$ degenerates to mutual information when $M = 2$. The $\sum_{i=1}^M \h(X^i)$ measures the total amount of information when treating $X^1,..., X^M$ independently; while the $\h(X^1,..., X^M)$ measures the counterpart when treating these $M$ variables as a whole. Therefore, the difference between them implies the redundant information, i.e., the information shared by these $M$ variables. 

Similar to mutual information, the $\tc$ is equivalent to the Kullback-Leibler ($\kl$)-divergence between $\mbox{P}_{\mathcal{X}^1 \times ... \times \mathcal{X}^M}$ and product of marginal distribution $\otimes_{m=1}^M \mbox{P}_{\mathcal{X}^m}$:
\begin{align}
    \label{eq:tc-kl}
    \tc(X^1,...,X^M) & = \mathrm{D}_{\kl}\left( d\mbox{P}_{\mathcal{X}^1 \times ... \times \mathcal{X}^M} \ || \  d\otimes_{m=1}^M \mbox{P}_{\mathcal{X}^m} \right) = \mathbb{E}_{\mbox{P}_{\mathcal{X}^1 \times ... \times \mathcal{X}^M}} \log{\frac{d\mbox{P}_{\mathcal{X}^1 \times ... \times \mathcal{X}^M}}{d\otimes_{m=1}^M \mbox{P}_{\mathcal{X}^m}} },
\end{align}
where $\mathrm{D}_{\kl}(\mathcal{P} \ || \ \mathcal{Q}) = \mathbb{E}_{\mathcal{P}}\log{\frac{d\mathcal{P}}{d\mathcal{Q}}}$. Intuitively, larger $\kl$ divergence between joint and marginal distribution indicates more dependence among these $M$ variables. To better characterize such a property, we give a formal definition of ``Point-wise Total Correlation" (PTC):
\begin{definition}[Point-wise Total Correlation]
\label{def:ptc}
Given $M$ random variables $X^1,...,X^M$, the Point-wise Total Correlation on $(x^1,...,x^m) \in \mathcal{X}^1 \times ... \times \mathcal{X}^M$, i.e., $\ptc(x^1,...,x^M)$ is defined as:
\begin{equation*}
    \ptc(x^1,...,x^M) = \log{ \frac{p(x^1,...,x^M)}{p(x^1)...p(x^M)} }
\end{equation*}
Further, the $R(x^1,...,x^M) := \frac{p(x^1,...,x^M)}{p(x^1)...p(x^M)}$ is denoted as the joint-margin ratio. 
\end{definition}

\begin{remark}
The Point-wise Total Correlation can be understood as the point-wise distance between joint distribution and the marginal distribution. In more details, as noted from \cite{huang2015euclidean}, by applying first-order Taylor-expansion, we have $\log{\frac{p(x)}{q(x)}} \approx \log{1} + \frac{p(x) - q(x)}{q(x)} = \frac{p(x) - q(x)}{q(x)}$. Therefore, the expected value of $\ptc(\cdot)$ can well measure the amount of information shared among these variables, which will be shown later in detailed.  
\end{remark}

For simplicity, we denote $p^{[M]}(x) := p(x^1,...,x^M)$ and $q^{[M]}(x) := \prod_{i=1}^M p(x^i)$. According to dual representation in \cite{nguyen2010estimating}, we have the following lower bound for $\kl$ divergence between $p$ and $q$, and hence $\tc$. 

\begin{lemma}[Dual version of $f$-divergence \cite{nguyen2010estimating}]
\label{lemma:dual-tc}
\begin{equation}
\label{eq:dv}
    \mathrm{D}_{\kl}\left( p^{[M]} \ || \  q^{[M]} \right) \geq \sup_{g \in \mathcal{G}} \mathbb{E}_{x \sim p^{[M]}} [g(x)] - \mathbb{E}_{x \sim q^{[M]}} \left[ e^{g(x) - 1} \right]
\end{equation}
where $\mathcal{G}$ is the set of functions that maps $\mathcal{X}^1\times\mathcal{X}^2 \times \dots\times \mathcal{X}^M$ to $\mathbb{R}$. The equality holds if and only if $g(x^1, x^2, \dots, x^M)  = 1 + \ptc(x^1,...,x^M)$, and the supremum is $\mathrm{D}_{\kl}\left( p^{[M]} \ || \  q^{[M]} \right) = \mathbb{E}_{p^{[M]}}(\ptc)$.
\end{lemma}
The Lemma~\ref{lemma:dual-tc} is commonly utilized for estimation of Mutual information \cite{belghazi2018mine} or optimization as variational lower bound in the machine learning literature. Besides, it also informs that the $\ptc$ is the optimal function to describe the amount of information shared by these $M$ variables. Indeed, such shared information is the information intersection among these variables, i.e., \emph{conditional on such information, these $M$ variables are independent of each other}. To quantitatively describe this, we first introduce the \emph{conditional total correlation} (CTC). Similar to TC, CTC measures the amount of information shared by these $M$ variables conditioning on some variable $Z$:
\begin{definition}[Conditional Total Correlation (CTC)]
\label{def:ctc}
Given $M+1$ random variables $X^1,...,X^M,Z$, the Conditional Total Correlation ($\ctc(X^1,...,X^M | Z)$) is defined as
\begin{equation*}
    \ctc(X^1,...,X^M | Z) = \sum_{i=1}^M \h(X^i | Z) - \h(X^1,...,X^M | Z)
\end{equation*}
\end{definition}

\section{Method}
\paragraph{Problem statement} In the semi-supervised muli-modal learning scenario, we have access to an unlabeled dataset $\mathcal{D}_u = \{x_i^{[M]}\}_i$ and a labeled dataset $\mathcal{D}_l = \{(x_i^{[M]}, y_i) \}_i$. Each label $y_i \in \mathcal{C}$, where $\mathcal{C}$ denotes the set of classes. Each datapoint $x_i^{[M]} := \{x_i^1, x_i^2, \dots, x_i^M | x_i^m \in \mathcal{X}^m \}$ consists of $M$ modalities, where $\mathcal{X}^m$ denotes the domain of the $m$-th modality. Datapoints and labels in $\mathcal{D}_l$ are i.i.d. samples drawn from the joint distribution $U_{X^{[M]}, Y}(x^1, x^2, \dots, x^M, y) := \Pr(X^1=x^1, X^2=x^2, \dots, X^M=x^M, Y=y)$. Data points in $\mathcal{D}_u$ are i.i.d. samples drawn from joint distribution $U_{X^{[M]}}(x^1, x^2, \dots, x^M) :=  \sum_{c \in \mathcal{C}} U_{X^{[M]}, Y}(x^1, x^2, \dots, x^M, y=c)$. Denote the prior of the ground truth labels by $\mathbf{p}^* = (\Pr(Y=c))_c$. Upon the labeled and unlabeled datasets, our goal is to train $M$ classifiers $h^{[M]} := \{h^1, h^2, \dots, h^M\}$ and an aggregator $\zeta$ such that $\forall m, h^m: \mathcal{X}^m \to \Delta_{\mathcal{C}}$ predicts the ground truth $y$ based on a $m$-th modality $x^m$ and $\zeta:\mathcal{X}^1\times\mathcal{X}^2\times\cdots\times\mathcal{X}^M \to \Delta_{\mathcal{C}}$ predicts the ground truth $y$ based on all of the modalities $x^{[M]}$. 

\paragraph{Outline} We will first introduce the assumptions regarding the ground truth label $Y$ and prior distribution on $(X^1,..., X^M, Y)$ in section~\ref{sec:model-assump}. In section~\ref{sec:pursue}, we will present our method, i.e., maximize the total correlation gain on unlabeled dataset $\mathcal{D}_u$. Finally, we will introduce our algorithm for optimization in section~\ref{sec:opt}.

\subsection{Assumptions for Identification of $Y$}
\label{sec:model-assump}

In this section, we first introduce two basic assumptions to ensure that the ground-truth label can be identified. According to Proposition 2.1 in \cite{achille2018emergence}, the label $Y$ can be viewed as the generating factor of data $X$. Such a result can be extended to multiple variables (please refer supplementary for details), which implies that $Y$ is the common generating factor of  $X^1,..., X^M$. Motivated by this, it is natural to assume that the ground truth label $Y$ is the "information intersection" among $X^1,..., X^M$, i.e., all of the modalities are independent conditioning on the ground-truth:  


\begin{assumption}[Conditional Independence]
\label{cond}
Conditioning on $Y$, $X^1, X^2, \dots, X^M$ are independent, i.e., $\forall x^1, \dots, x^M$, 
\begin{align*}
    \Pr(X^{[M]}=x^{[M]} | Y=c) = \prod_{m} \Pr(X^m=x^m | Y=c), \ \mathrm{for \ any } c \in \mathcal{C}.
\end{align*}
\end{assumption}

On the basis of this assumption, one can immediately get the conditional total correlation gain $\ctc(X^1,...,X^M | Y) = 0$. In other words, conditioning on $Y$, there is no extra information shared by these $M$ modalities, which is commonly assumed in the literature of semi-supervised learning \cite{lewis1998naive,blum1998combining,dasgupta2002pac,leskes2005value}. However, the $Y$ may not be the unique information intersection among these $M$ modalities. Specifically, the following lemma establishes the rules for such information intersection to hold:

\begin{lemma}
\label{lemma:PTC}
Given assumption~\ref{cond}, $R(x^1,...,x^M)$ (Joint-marginal ratio definition~\ref{def:ptc}) has
\begin{align*}
    R(x^1,...,x^M) = \sum_{c \in \mathcal{C}} \frac{ \prod_{m} \Pr(Y=c | X^m=x^m)}{ (\Pr(Y=c))^{M-1}}
\end{align*}
Further, the optimal $g$ in lemma \ref{lemma:dual-tc} satisfies $g(x^1,...,x^M) = 1 + $ $\log{\sum_{c \in \mathcal{C}} \frac{ \prod_{m} \Pr(Y=c | x^m)}{ (\Pr(Y=c))^{M-1}}}$.
\end{lemma}

In other words, in addition to $\{\Pr(Y=c | X^m=x^m)\}_{m}, \mathbf{p}^*_c$, there are other solutions $\{a_{x^1},...,a_{x^M}\},r$ with $a_{x^i} \in \Delta_{\mathcal{C}}$ (for $i \in \mathcal{C}$) and $r \in \Delta_{\mathcal{C}}$ that can make the $g$ optimal, as long as its joint-marginal ratio is equal to the ground-truth one:
\begin{equation}
\label{eq:prior-condition}
    \sum_{c \in \mathcal{C}} \frac{ \prod_{m} a_{x^m}}{ (r^a_c)^{M-1}} = R(x^1,...,x^M)
\end{equation}
 To make $\{\Pr(Y=c | X^m=x^m)\}_{m}, \mathbf{p}^*_c$ identifiable w.r.t a trivial permutation, we make the following trivial assumption on $\Pr(X^1,...,X^M,Y)$.

\begin{assumption}[Well-defined Prior]
\label{prior-well}
The solutions $\{a_{x^1},...,a_{x^M}\},r^a$ and $\{b_{x^1},...,b_{x^M}\},r^b$ for Eq.~\eqref{eq:prior-condition} are equivalent under the permutation $\prod: \mathcal{C} \to \mathcal{C}$: $ a_{x^m} = \prod b_{x^m}, \ r^a = \prod r^b$.
\end{assumption}


\subsection{Total Correlation Gain Maximization (TCGM)}
\label{sec:pursue}

Assumption~\ref{cond} indicates that the label $Y$ is the generating factor of all modalities, and assumption~\ref{prior-well} further ensures its uniqueness under permutation. Our goal is to learn the ground-truth label $Y$ which is the information intersection among $M$ modalities. In this section, we propose a novel framework, namely \emph{\textbf{Total Correlation Gain Maximization}} (\textbf{TCGM}) to capture such an information intersection, which is illustrated in Figure.~\ref{fig:overview}. To the best of our knowledge, we are the first to theoretically prove the identification of ground truth classifiers on semi-supervised multi-modality data, by generalizing \cite{kong2018water,cao2018max} that can only handle two views in multi-view scenario. The high-level spirit is designing TC-induced loss over classifiers of every modality. By maximizing such a loss, these classifiers can converge to Bayesian posterior, which is the optimal solution of TC as expectation of the loss. First, we introduce the basic building blocks for our method.

\paragraph{Classifiers $h^{[M]}$}
In order to leverage the powerful representation ability of deep neural network (DNN), each classifier $h^m(x^m;\Theta^m)$ is modeled by a DNN with parameters $\Theta^m$. For each modality $m$, we denote the set of all such classifiers by $H^m$ and $H^{[M]} := \{H^1, H^2, \dots, H^M\}$.

\paragraph{Modality Classifiers-Aggregator $\zeta$} Given $M$ classifiers for each modality $h^{[M]}$ and a distribution $\mathbf{p}=(p_c)_c \in \Delta_\mathcal{C}$, the aggregator $\zeta$ which predicts the ground-truth label by aggregating classifiers of all modalities, is constructed by \[\zeta(x^{[M]}; h^{[M]}, \mathbf{p}) = \text{Normalize}\left( \left(\frac{\prod_m h^m(x^m_i)_c}{(p_c)^{M-1}}\right)_c \right) \]
where $\text{Normalize}(\mathbf{v}) := \frac{\mathbf{v}}{\sum_c v_c}$ for all $\mathbf{v} \in \Delta_{\mathcal{C}}$. 
\paragraph{Reward Function $\mathcal{R}$} We define a reward function $\mathcal{R}: \overbrace{\Delta_{\mathcal{C}} \times ... \times \Delta_{\mathcal{C}}}^{M} \to \mathbb{R}$ that measures the "amount of agreement" among these classifiers.
Note that the desired classifiers should satisfy Eq.~\eqref{eq:prior-condition}. \newline

Inspired by Lemma~\ref{lemma:dual-tc}, we can take the empirical total correlation gain of $N$ samples, i.e., the lower bound of Total Correlation as our maximization function. Specifically, given a reward function $\mathcal{R}$, the empirical total correlation with respect to classifiers $h^{[M]}$, a prior $\mathbf{p} \in \Delta_{\mathcal{C}}$ measures the empirical "amount of agreement" for these $M$ classifiers at the same sample $(x^1_i,...,x^M_i) \in \mathcal{D}_u$, with additionally a punishment of them on different samples: $x^1_{i_1} \in \mathcal{X}^1,...,x^M_{i_M} \in \mathcal{X}^M$ with $i_1 \neq i_2 \neq ... \neq i_M$:
\begin{align}
    \label{eq:emp-tc}
     TCg[\mathcal{R}](\{x_i^{[M]}\}_{i=1}^N; h^{[M]},p) & := \frac{1}{N}\sum_{i} \mathcal{R}(h^1(x^1_i),...,h^M(x^M_i)) \nonumber \\
    & \ \ \ \ \ \ \ \ \ \ \ \ \  - \frac{1}{N!/(N-M)!}\sum_{i_1\neq i_2 \neq \cdots \neq i_M} e^{\mathcal{R}(h^1(x^1_{i_1}),...,h^M(x^M_{i_M})) - 1}
\end{align}
for simplicity we denote $TCg[\mathcal{R}](\{x_i^{[M]}\}_{i=1}^N; h^{[M]},p)$ as $TCg^{(N)}$. Intuitively, we expect our classifiers to be consistent on the same sample; on the other hand, to disagree on different samples to avoid learning a trivial solution that classifies all data points into the same class. 

\begin{definition}
[Bayesian posterior classifiers/aggregator] The $h_{*}^{[M]}$ and $\zeta_{*}$ are called \emph{Bayesian posterior classifiers} and \emph{Bayesian posterior aggregator} if they satisfy
\begin{align*}
    \forall m, h_{*}^{m}(x^m)_c = \Pr(Y=c | X^m = x^m); \ \zeta_{*}(x^{[M]})_c = \Pr(Y=c | X^{[M]} = x^{[M]}).
\end{align*}
\end{definition}

Note that from Eq.~\eqref{eq:emp-tc} that our maximization goal, i.e., $TCg^{(N)}$ relies on the form of reward function $\mathcal{R}$. The following Lemma tells us the form of optimal reward function, with which we can finally give an explicit form of $TCg^{(N)}$. 
\begin{lemma}
\label{lemma:optimal-reward}
The $\mathcal{R}_{*}$ that maximizes the expectation of $TCg^{(N)}$ can be represented as the Point-wise Total Correlation function, which is the function of Bayesian classifiers and the prior of ground truth labels $(\mathbf{p}^*_c)_c$:
\begin{align}
    \mathcal{R}_{*}(h^1(x^1),...,h^M(x^M)) = 1 + \ptc(x^1,...,x^M) 
                                        = 1 + \log{\sum_{c \in \mathcal{C}} \frac{\prod_m h^m_{*}(x^m)_c}{(\mathbf{p}^*_c)^{M-1}}} \nonumber
\end{align}
\end{lemma}

\paragraph{Total Correlation Gain} Bring $\mathcal{R}_{*}$ to Eq.~\eqref{eq:emp-tc}, we have:
\begin{align}
    \label{eq:empi}
     TCg(\{x_i^{[M]}\}_{i=1}^N; h^{[M]},p) := & 1 + \frac{1}{N}\sum_{i} \log{\sum_{c \in \mathcal{C}} \frac{ \prod_{m} h^m(x^m_i)_c }{ (p_c)^{M-1} } }  \nonumber \\
    - &\frac{1}{N!/(N-M)!}\sum_{i_1\neq i_2 \neq \cdots \neq i_M} \sum_{c \in \mathcal{C}} \frac{ \prod_{m} h^m(x^m_{i_m})_c }{ (p_c)^{M-1} }.
\end{align}
As inspired by Lemma~\ref{lemma:optimal-reward}, we have that these Bayesian posterior classifiers are maximizers of the expected total correlation gain. Therefore, we can identify the equivalent class of Bayesian posteriors by minimizing $-TCg^{(N)}$ on unlabeled dataset $\mathcal{D}_u$. By additionally minimize expected cross entropy (CE) loss on $\mathcal{D}_l$, we can identify the unique Bayesian classifiers since they are respectively the minimizers of CE loss.


\begin{theorem}[Main theorem] 
Define the expected total correlation gain $eTCg(h^{[M]},p)$:
\begin{equation*}
    eTCg(h^1,...,h^M,p) := \mathbb{E}_{x_i^{[M]}\sim U_{X^{[M]}}} \left(TCg(x_i^{[M]}; h^{[M]},p) \right)
\end{equation*}
Given the conditional independence assumption~\ref{cond} and well-defined prior assumption~\ref{prior-well}, we have that the maximum value of $eTCg$ is Total Correlation of $M$ modalities, i.e., $\mathrm{TC}(X^1,...,X^M)$. Besides,
\paragraph{Ground-truth $\to$ Maximizer} $(h_{*}^{[M]}, \mathbf{p}^{*})$ is a maximizer of $eTCg(h^{[M]},p)$. In other words, $\forall h^{[M]}\in H^{[M]}, p \in \Delta_{ \mathcal{C}}$, $eTCg(h_*^1,...,h_*^M,\mathbf{p}^{*}) \geq eTCg(h^1,...,h^M,p)$. The corresponding optimal aggregator is $\zeta_{\star}$, i.e., $\zeta_{\star}(x^{[M]})_c = \mathrm{Pr}(Y = c | X^{[M]} = x^{[M]})$.
\paragraph{Maximizer $\to$ (Permuted) Ground-truth } If the prior is well defined, then for any maximizer of $eTCg$, $(\tilde{h}^{[M]}, \tilde{\mathbf{p}})$, there is a permutation $\tilde{\prod}: \mathcal{C} \to \mathcal{C}$ such that:
\begin{align}
\label{eq:aggregator}
    \tilde{h}^m(x^m)_c = \mathrm{P}(Y = \tilde{\prod}(c) | X^m = x^m), \  \tilde{\mathbf{p}}_c = \mathrm{P}(Y = \tilde{\prod}(c))
\end{align}
\end{theorem}
The proof is in Appendix A. Note from our main theorem that by maximizing the $eTCg$, we can get the total correlation of $M$ modalities, which is the ground-truth label $Y$, and also the equivalent class of Bayesian posterior classifier under permutation function. In order to identify the Bayesian posterior classifiers, we can further minimize cross-entropy loss on labeled data $\mathcal{D}_l$ since the Bayesian posterior classifiers are the only minimizers of the expected cross-entropy loss. On the other hand, compared with only using $\mathcal{D}_l$ to train classifiers, our method can leverage more information from $\mathcal{D}_u$, which can be shown in the experimental result later.



\subsection{Optimization}
\label{sec:opt}

\begin{figure*}[h!]
    \centering
    \includegraphics[width=0.9\textwidth]{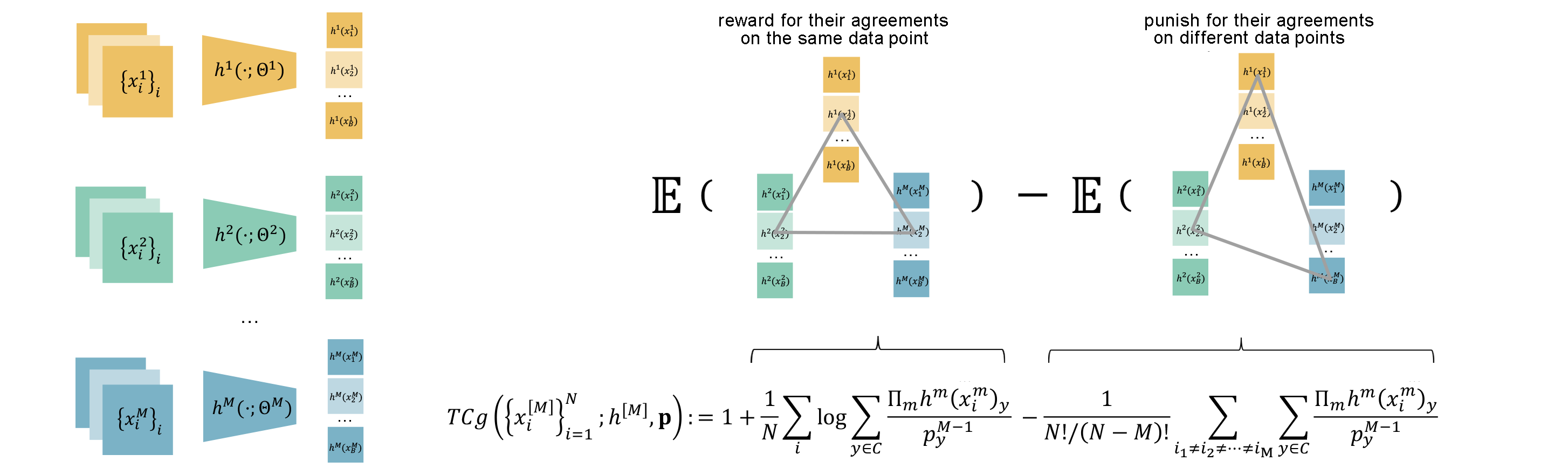}
    \caption{Empirical Total Correlation Gain $TCg^{(N)}$}
    \label{fig:overview}
\end{figure*}

Since $eTCg$ is intractable, we alternatively maximize the empirical total correlation gain, i.e., $TCg^{(N)}$ to learn the optimal classifiers. To identify the unique Bayesian posteriors, we should further utilize labeled dataset $\mathcal{D}_l$ in a supervised way. Our whole optimization process is shown in Appendix, which adopts iteratively optimization strategy that is roughly contains two steps in each round: (i) We train the $M$ classifiers using the classic cross entropy loss on the labeled dataset $\mathcal{D}_l$ and (ii) using our information-theoretic loss function $\mathcal{L}_{\text{TC}}$ on the unlabeled dataset $\mathcal{D}_u$. To learn the Bayesian posterior classifiers more accurately, the (ii) can help to learn the equivalent class of Bayesian Posterior Classifiers and (i) is to learn the correct and unique classifiers. As shown in Figure \ref{fig:overview}, by optimizing $\mathcal{L}_{\text{TC}}^{(B)}$ (Eq.~\eqref{eq:emp-tc} with $B$ denoting the number of samples in each batch), we reward the $M$ classifiers for their agreements on the same data point and punish the $M$ classifiers for their agreements on different data points. 

\paragraph{Loss function $\mathcal{L}_{\text{CE}}$ for labeled data} We use the classic cross entropy loss for labeled data. Formally, for a batch of data points $\{x_i^{[M]}\}_{i=1}^B$ drawn from labeled data $\mathcal{D}_{l}$, the cross entropy loss $\mathcal{L}_{\text{CE}}$ for each classifier $h^m$ is defined as $\mathcal{L}_{\text{CE}}(\{(x_i^{[M]},y_i)\}_{i=1}^B; h^{m}) :=\frac{1}{B}\sum_i -\log(h^m(x^m_i)_{y_i})$.

\paragraph{Loss function $\mathcal{L}_{\text{TC}}^{(B)}$ for unlabeled data} For a batch of data points $\{x_i^{[M]}\}_{i=1}^B$ drawn from unlabeled data $\mathcal{D}_{u}$, our loss function $\mathcal{L}_{\text{TC}}^{(B)} := -TCg^{(B)}$ that is defined in Eq.~\eqref{eq:empi} with $N$ replaced by number of batch size $B$. When $N$ is large, we only sample a fixed number of samples from product of marginal distribution to estimate the second term in Eq.~\eqref{eq:empi}, which makes training more amenable.

\paragraph{Prediction} After optimization, we can get the classifiers $\{h_m\}_m$. The prior $\mathbf{p}_c$ can be estimated from data, i.e., $\mathbf{p}_c = \frac{\sum_{i \in [N]}\mathbf{1}(y_i = c)}{N}$. Then based on Eq.~\eqref{eq:aggregator}, we can get the aggregator classifier $\zeta$ for prediction. Specifically, given a new sample $\tilde{x}^{[M]}$, the predicted label is $\tilde{y} := \arg\max_{c} \zeta(\tilde{x}^{[M]})_c$.

\paragraph{Time Complexity} The overall time complexity for optimizing our TCGM is linear scale to the number of modalities, i.e., $\mathcal{O}(M)$. Please refer to appendix for detailed analysis. 

\section{Preliminary experiments}

\begin{wrapfigure}{r}{0.4\textwidth}
    \includegraphics[width=0.4\textwidth]{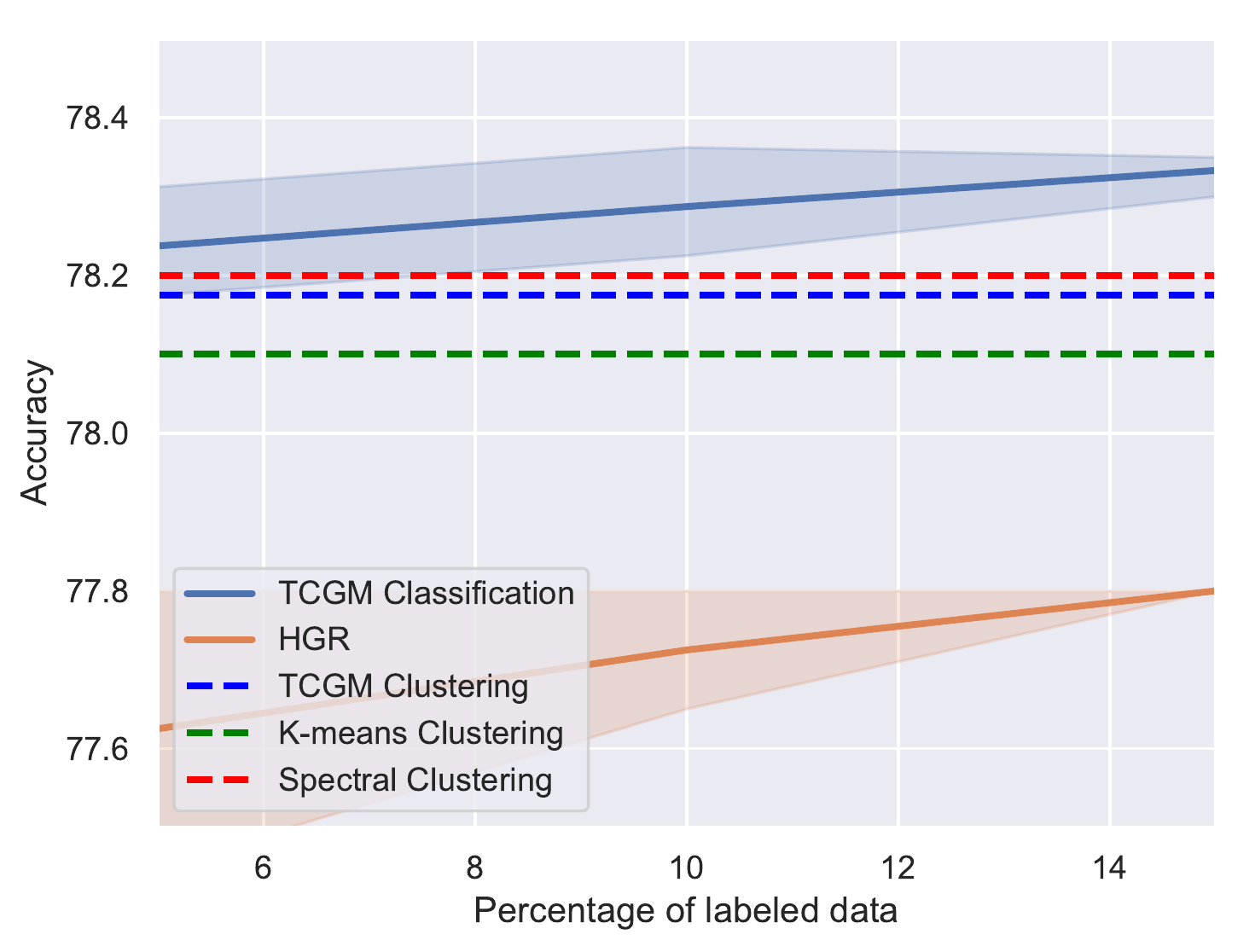}
    \caption{Clustering and classification accuracy.}
  \label{fig:acc}
\end{wrapfigure}
We first conduct a simulated experiment on synthetic data to validate our theoretical result of \textbf{TCGM}. Specifically, we will show the effectiveness of Total Correlation Gain $TCg$ for unsupervised clustering of data. Further, with few labeled data, our \textbf{TCGM} can give accurate classification. In more detail, we synthesize the data of three modalities from a specific Gaussian distribution $P(X^i|y) \ (i=1,2,3)$. The clustering accuracy is calculated as classification accuracy by assuming the label is known. As shown in Figure~\ref{fig:acc}, our method \textbf{TCGM} has competitive performance compared to well established clustering algorithms K-means++ \cite{ashburner2007fast} and spectral clustering \cite{Ng2001OnSC}. Based on the promising unsupervised learning result, as shown by the light blue line (the top line) in Figure~\ref{fig:acc}, our method can accurately classify the data even with only a small portion of labeled data. In contrast, \textbf{HGR}~\cite{wang2018efficient} degrades since it can only capture the linear dependence and fail when higher-order dependency exists. 

\section{Applications}
In this section, we evaluate our method on various multi-modal classification tasks: (i) News classification (Newsgroup) (ii) Emotion recognition: \textbf{IEMOCAP}, \textbf{MOSI} and (iii) Disease prediction of Alzheimer's Disease on 3D medical Imaging: \textbf{Alzheimer’s Disease Neuroimaging Initiative (ADNI)}. Our method \textbf{TCGM} is compared with : \textbf{CE} separately trains classifiers for each modality by minimizing cross entropy loss of only labeled data; \textbf{HGR} \cite{wang2018efficient} learns representation by maximizing correlation of different modalities; and \textbf{LMF}~\cite{Liu2018EfficientLM} performs multimodal fusion using low-rank tensors. 
The optimal hyperparameters are selected according to validation accuracy, among which the learning rate is optimized from $\{0.1, 0.01, 0.001, 0.0001\}$. All experiments are repeated five times with different random seeds. The mean test accuracies and standard deviations of single classifiers the aggregator ($\zeta$) are reported.


\subsection{News Classification}
\paragraph{Dataset} \textbf{Newsgroup} \cite{bisson2012co} \footnote{\url{http://qwone.com/~jason/20Newsgroups/}} is a group of news classification datasets. Following \cite{hussain2010icmla}, each data point has three modalities, PAM, SMI and UMI, collected from three different preprocessing steps \footnote{PAM (Partitioning Around Medoïds preprocessing), SMI (Supervised Mutual Information preprocessing) and UMI (Unsupervised Mutual Information preprocessing)}. We evaluate \textbf{TCGM} and the baseline methods on $3$ datasets from \textbf{Newsgroup}: News-M2, News-M5, News-M10. They contain 500, 500, 1000 data points with 2, 5, 10 categories respectively. Following \cite{yang2018semi}, we use $60\%$ for training, $20\%$ for validation and $20\%$ for testing for all of these three datasets.

\paragraph{Implementation details} We synthesize two different label rates (the percentage of labeled data points in each modality): $\{10\%, 30\%\}$ for each dataset. We follow \cite{yang2018semi} for classifiers. Adam with default parameters and learning rate $\gamma_u = 0.0001, \gamma_l = 0.01$ is used as the optimizer during training. Batch size is set to $32$. We further compare with two additional baselines: \textbf{VAT} \cite{vat} uses adversarial training for semi-supervised learning; \textbf{PVCC} \cite{yang2018semi} that considers the consistency of data points under different modalities.

\begin{figure*}[h!]
     \centering
     \includegraphics[width=\textwidth]{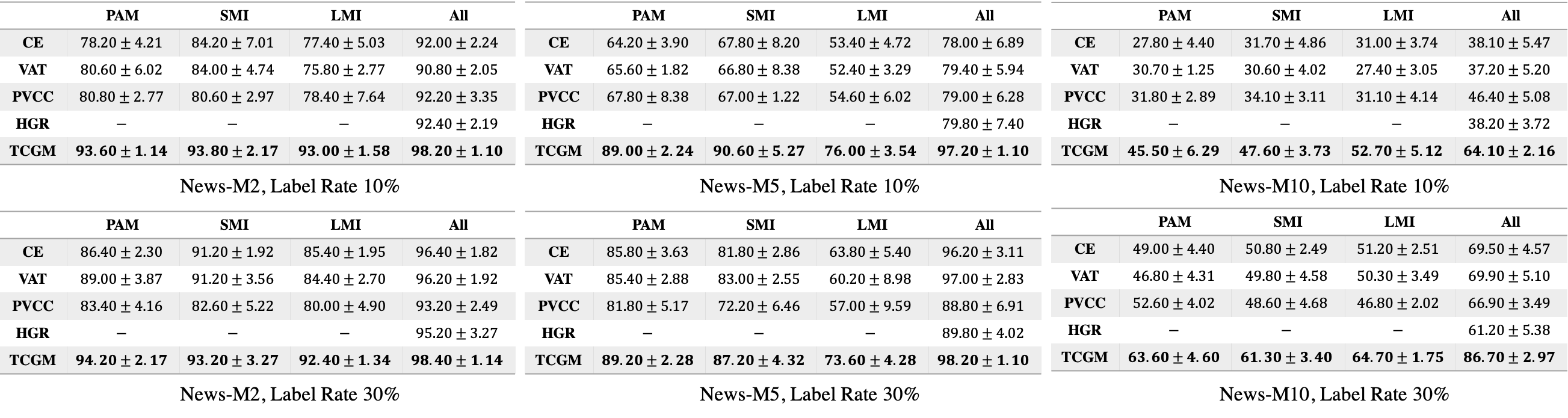}
     \caption{Test accuracies (mean $\pm$ std. dev. ) on Newsgroups datasets}
     \label{fig:news}
 \end{figure*}


 As shown in Fig.~\ref{fig:news}, \textbf{TCGM} achieves the best classification accuracy for both single classifier and aggregators, especially when the label rate is small. This shows the efficacy of utilizing the cross-modal information during training as compared to others that are unable to utilize the cross-modal information. Moreover, we can achieve further improvement by aggregating classifiers on all modalities, which shows the benefit of aggregating knowledge from different modalities.



\subsection{Emotion Recognition}
\paragraph{Dataset} We evaluate our methods on two multi-modal emotion recognition datasets: \textbf{IEMOCAP} dataset~\cite{Busso2008IEMOCAPIE} and \textbf{MOSI} dataset~\cite{Zadeh2016MOSIMC}. The goal for both datasets is to identify speaker emotions based on the collected videos, audios and texts. The IEMOCAP consists of 151 sessions of recorded dialogues, with 2 speakers per session for a total of 302 videos across the dataset. The MOSI is composed of 93 opinion videos from YouTube movie reviews. We follow the settings in \cite{Liu2018EfficientLM} for the data splits of training, validation and test set. For IEMOCAP, we conduct experiments on three different emotions: happy, angry and neutral emotions; for MOSI dataset we consider the binary classification of emotions: positive and negative. 

\paragraph{Implementation details} We synthesize three label rates for each dataset (the percentage of labeled data points in each modality): $\{0.5\%, 1\%, 1.5\%\}$ for IEMOCAP and $\{1\%, 2\%, 3\%\}$ for MOSI. For a fair comparison, we follow architecture setting in \cite{Liu2018EfficientLM}. We adopt the modality encoder architectures in \cite{Liu2018EfficientLM} as the single modality classifiers for CE and TCGM, while adopting the aggregator on the top of modality encoders for LMF and HGR. Adam with default parameters and learning rate $\gamma_u = 0.0001, \gamma_l = 0.001$ is used as the optimizer. The batch size is set to $32$.

\begin{figure}[h!]
    \centering
    \subfigure[Happy]{\includegraphics[width=0.3\textwidth]{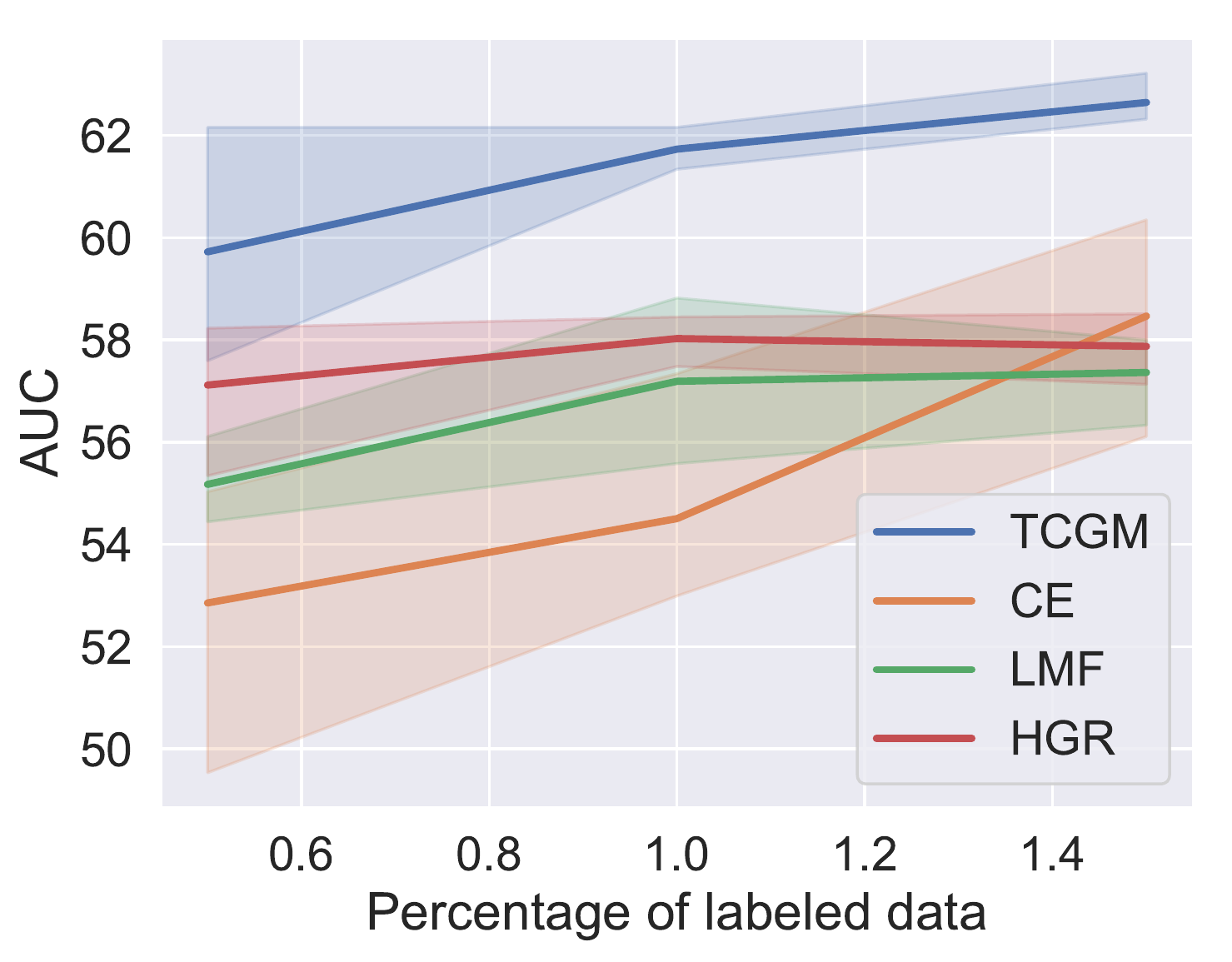}}
    \subfigure[Angry]{\includegraphics[width=0.3\textwidth]{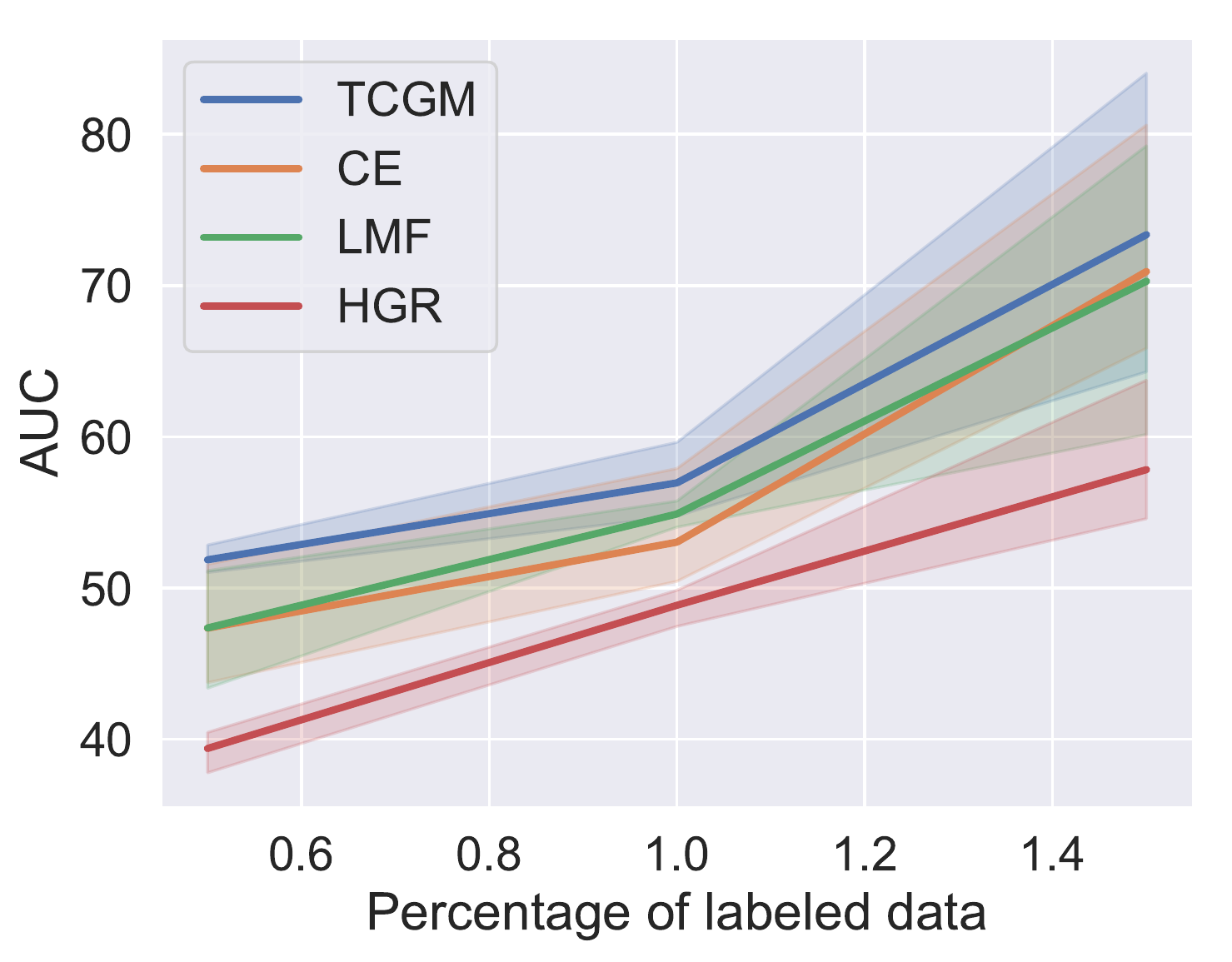}}
    \subfigure[Neutral]{\includegraphics[width=0.3\textwidth]{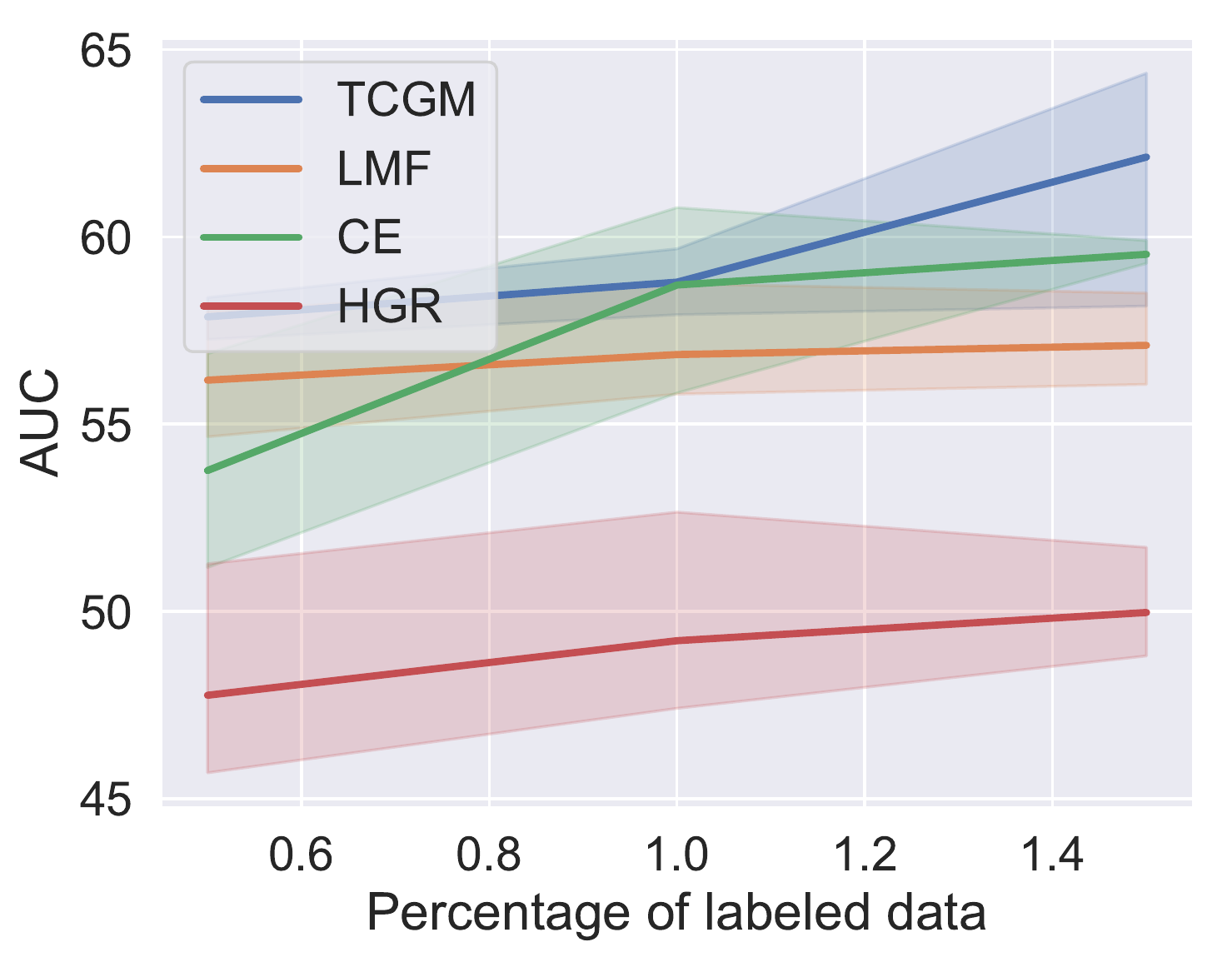}}
    \subfigure[Happy (text)]{\includegraphics[width=0.3\textwidth]{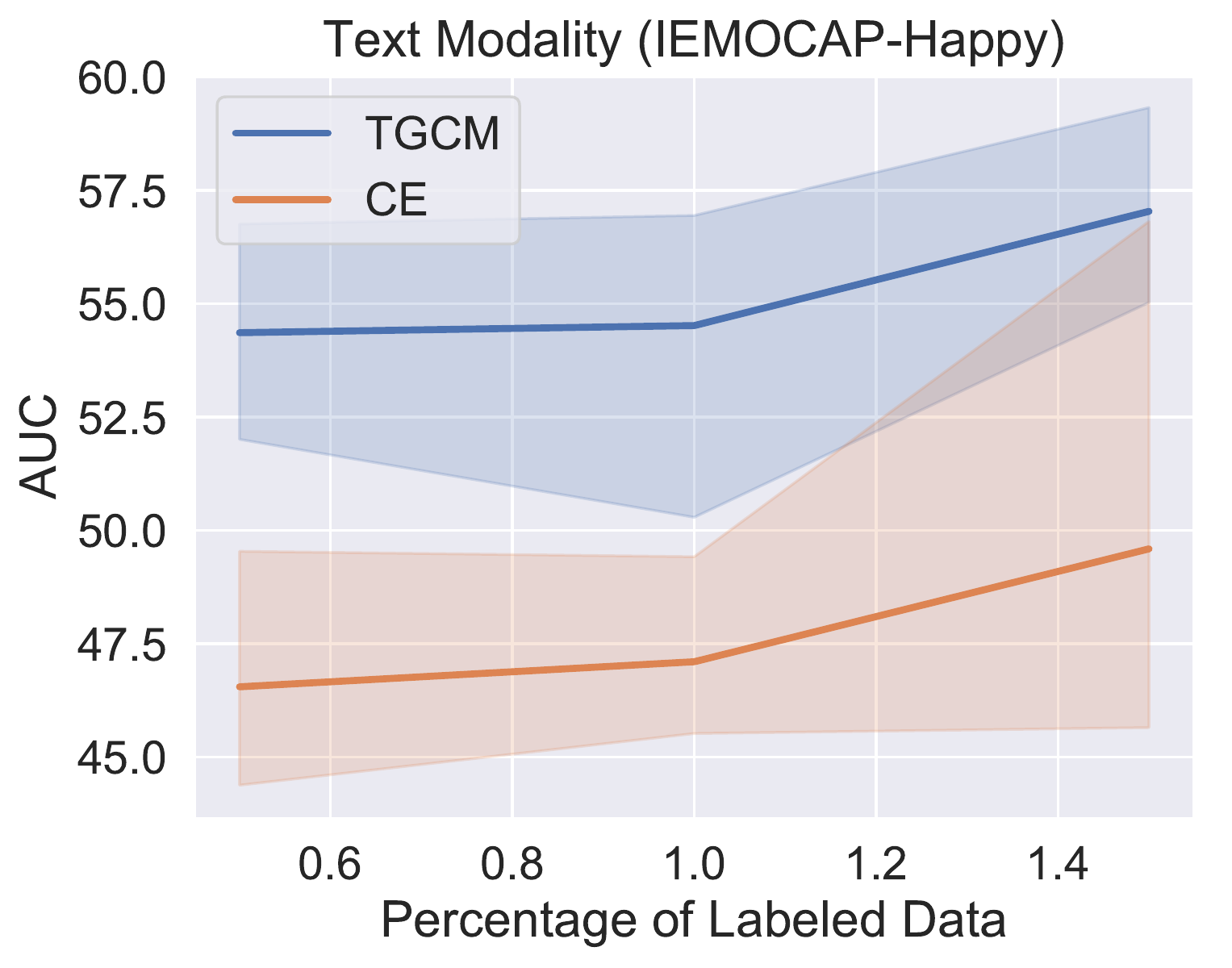}}
    \subfigure[Neutral (audio)]{\includegraphics[width=0.3\textwidth]{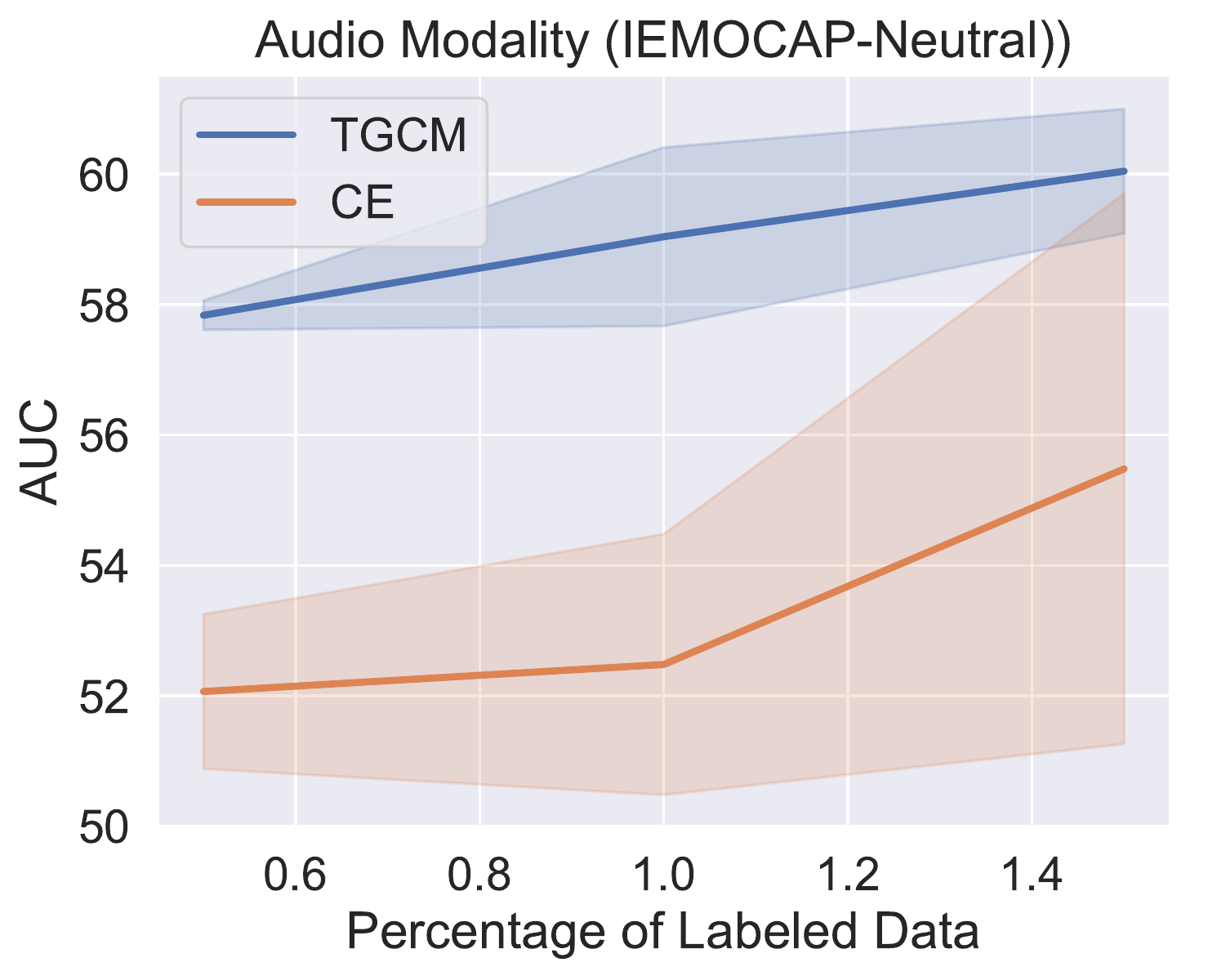}}
    \subfigure[Angry (video)]{\includegraphics[width=0.3\textwidth]{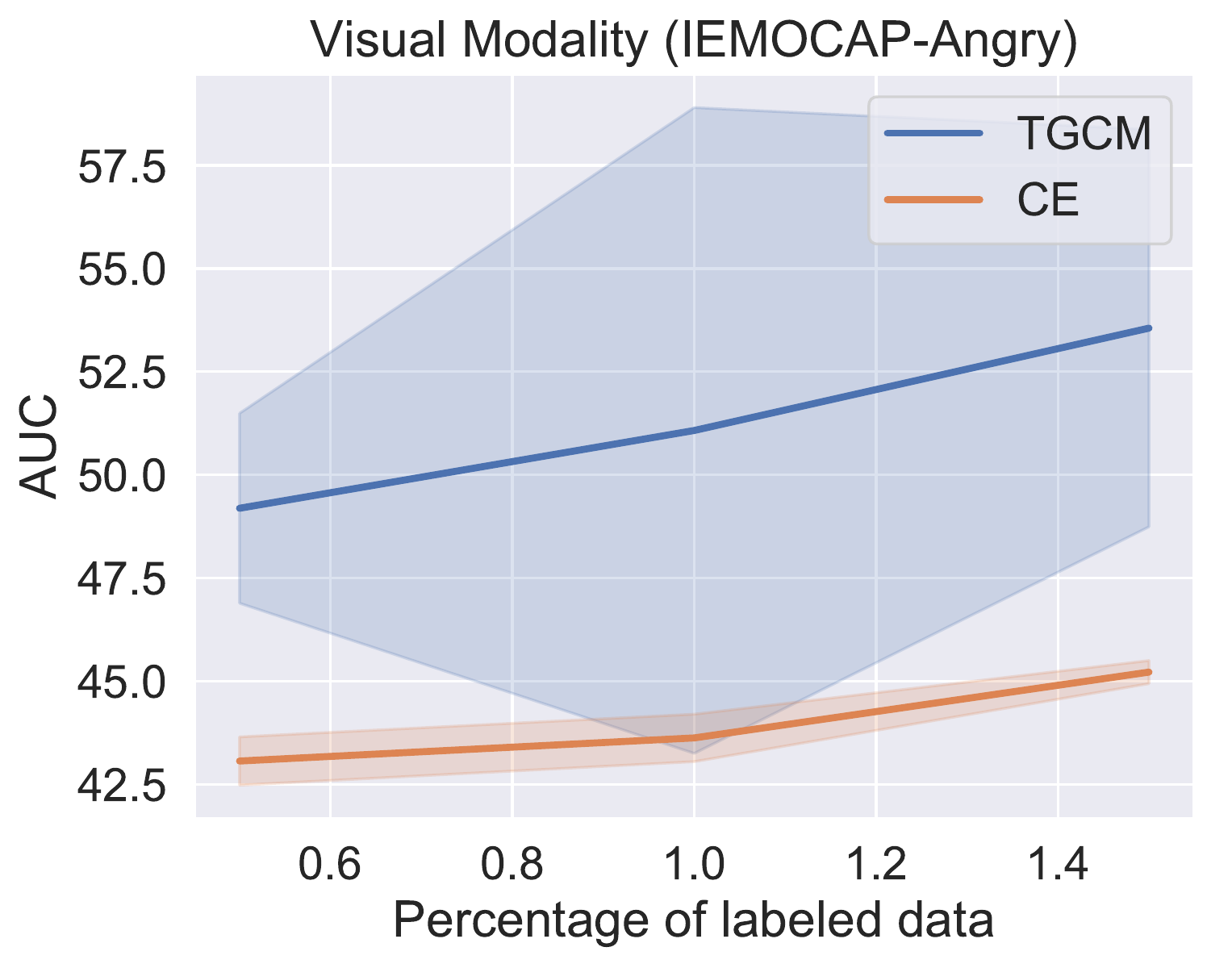}}
    \subfigure[MOSI]{\includegraphics[width=0.3\textwidth]{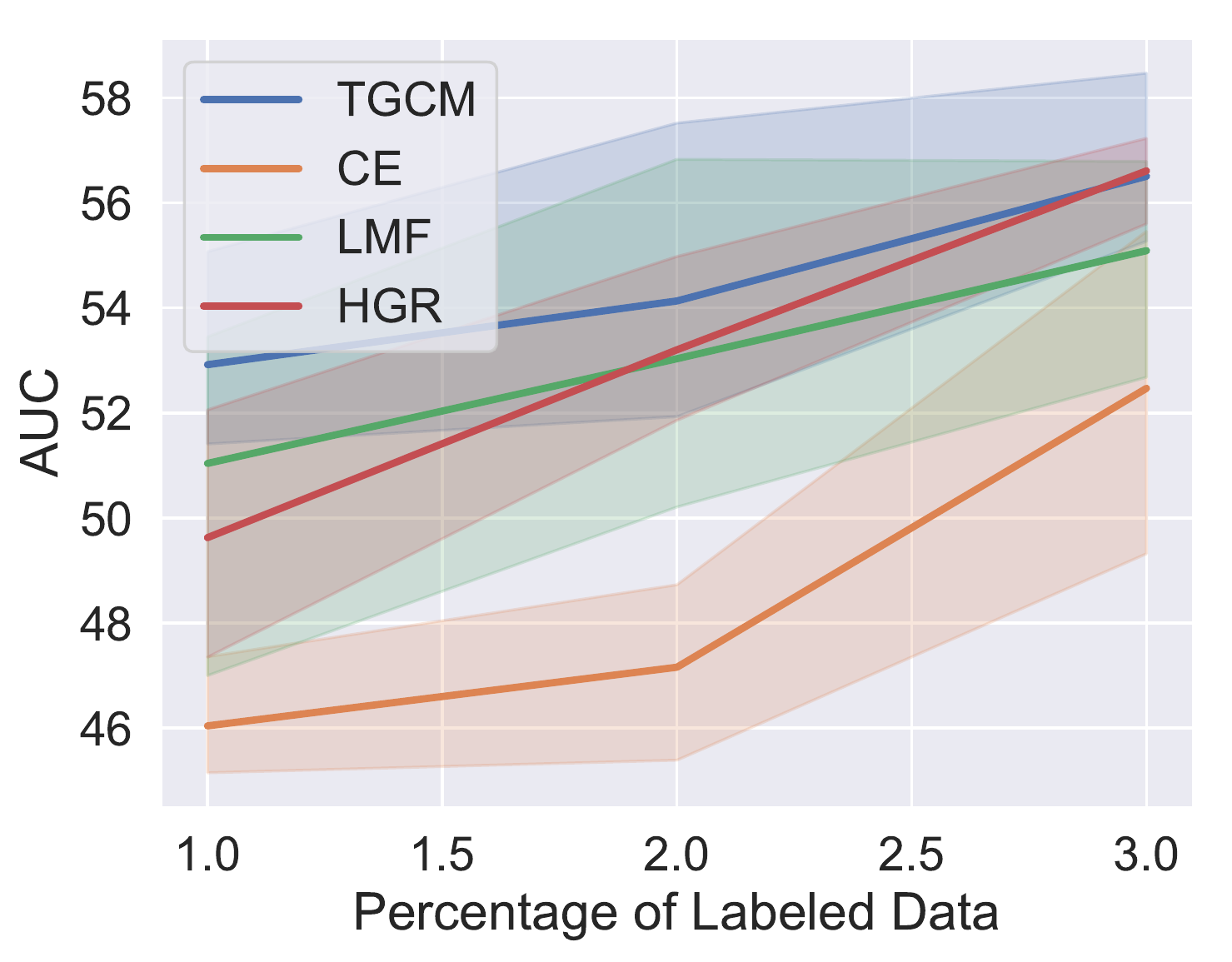}}
    \subfigure[MOSI (video)]{\includegraphics[width=0.3\textwidth]{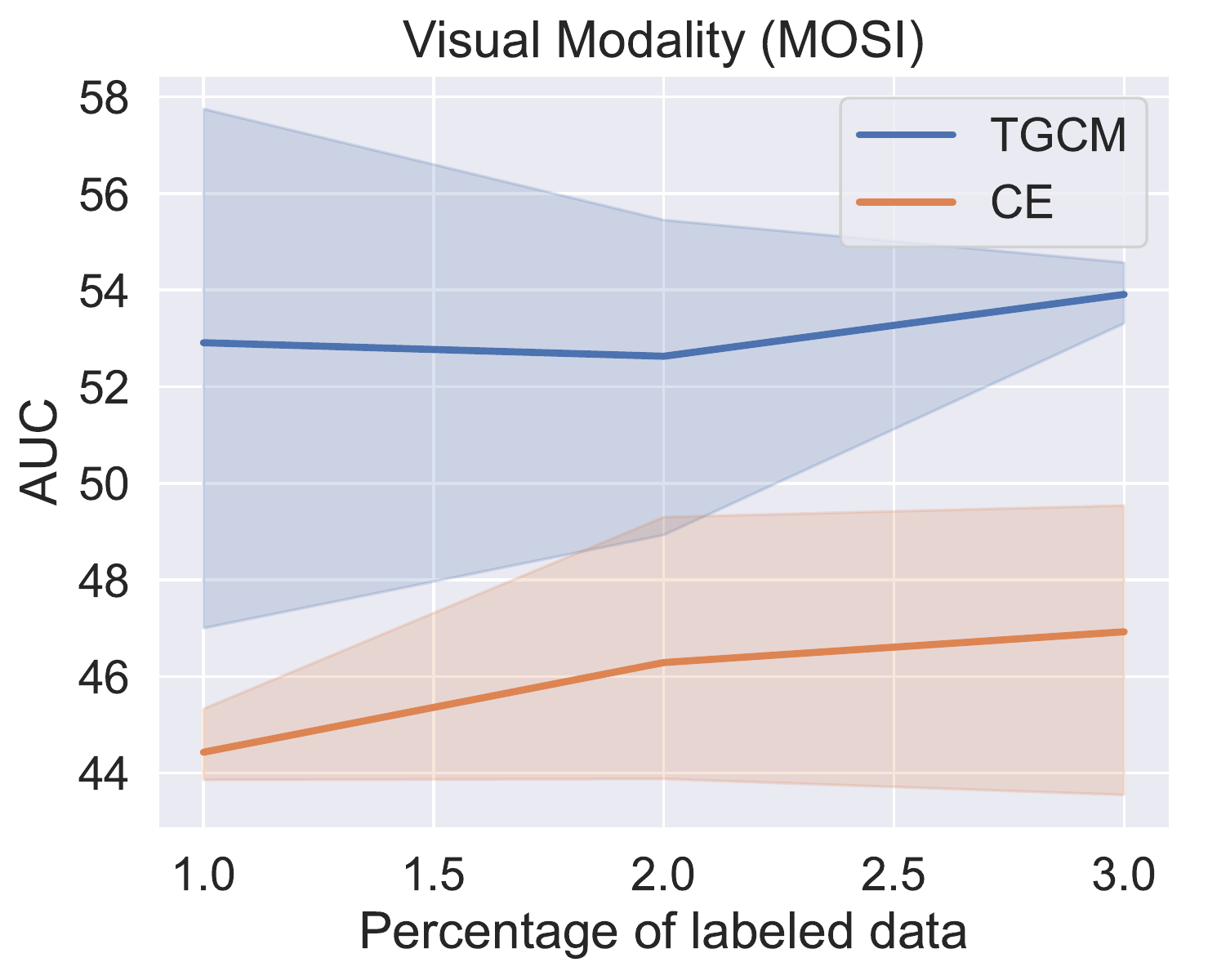}}
    \subfigure[MOSI
    (audio)]{\includegraphics[width=0.3\textwidth]{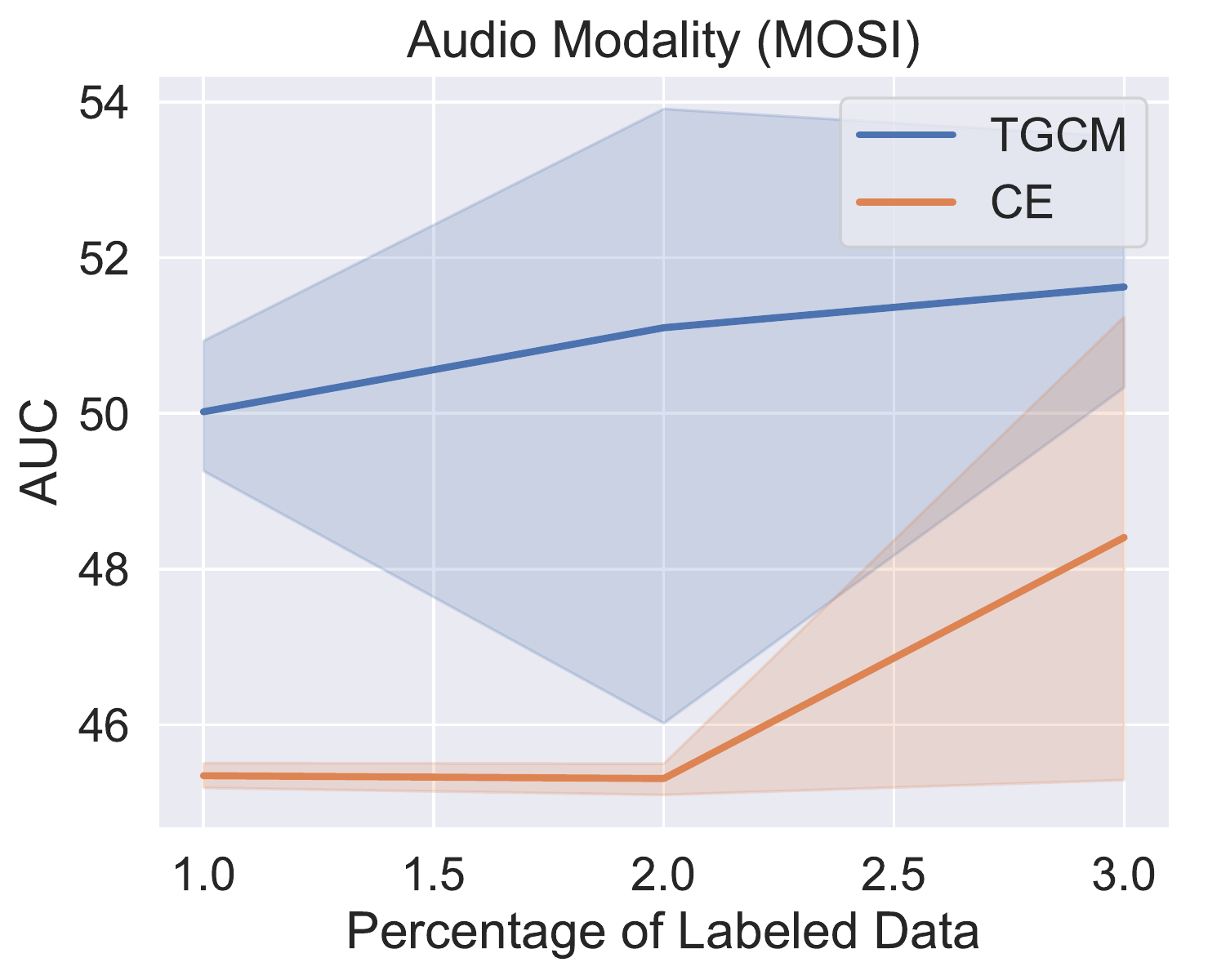}}
    \caption{\textbf{(a,b,c) AUC of Aggregators on happy, angry and neutral emotion recognition on IEMOCAP data:} \textbf{(d, e,f) AUC on text, audio and video modality modality classifiers on IEMOCAP:}  AUC of other composition of (modality, emotion) are listed in supplementary material. \textbf{(g,h,i) AUC on MOSI data}: AUC of (g) Aggregators on all modalities and single classifiers on (h) Video modality (i) Audio modality.}
    \label{fig:iemo}
\end{figure}

We report the AUC (Area under ROC curve) for the aggregators on all the modalities and single modality classifiers by different methods. We only report the AUC of LMF and HGR on all modalities since they do not have single modality classifiers. For single modality classifiers, we show results on the text modality on happy emotion (d), audio modality on neutral emotion (e) the video modality on angry emotion on IEMOCAP; and (h) the video modality (i) the audio modality on MOSI. Please refer to supplementary material for complete experimental result. As shown in Figure~\ref{fig:iemo}, aggregators trained by TCGM outperform all the baselines given only tiny fractions of labeled data. \textbf{TCGM} improves the AUC of the single modality classifiers significantly, which shows the efficacy of utilizing the cross-modal information during the training of our method. As label rates continue to grow, the advantage of our method over CE decreases since more information is provided for CE to learn the ground-truth label.

Our method also outperforms other methods in terms of the prediction based on all the modalities, especially when the label rate is small. This shows the superiority of our method when dealing with a limited amount of annotations.

\subsection{Disease prediction of Alzheimer's Disease}

\paragraph{Dataset} Early prediction of Alzheimer's Disease (AD) is attracting increasing attention since it is irreversible and very challenging. Besides, due to privacy issues and high collecting costs, an efficient classifier with limited labeled data is desired. To validate the effectiveness of our method on this challenging task, we only keep labels of a limited percentage of data, which is obtained from the most popular \textbf{Alzheimer’s Disease
Neuroimaging Initiative (ADNI)} dataset\footnote{\url{www.loni.ucla.edu/ADNI}}, with 3D images sMRI and PET. DARTEL VBM pipeline \cite{ashburner2007fast} is implemented to pre-process the sMRI data, and then images of PET were reoriented into a standard $91\times109\times91$ voxel image grid in MNI152 space, which is same with sMRIs'. To limit the size of images, only the hippocampus on both sides are extracted as input in the experiments. We denote subjects that convert to Alzheimer's disease (MCI$_c$) as AD, and subjects that remain stable (MCI$_s$) as NC (Normal Control). Our dataset contains 300 samples in total, with 144 AD and 156 NC. We randomly choose $60\%$ for training, $20\%$ for validation and $20\%$ for testing stage.

\paragraph{Implementation details} We synthesize two different label rates (the percentage of labeled data points): $\{10\%, 50\%\}$. DenseNet is used as the classifier. Two 3D convolutional layers with the kernel size $3\times 3\times 3$ are adopted to replace the first 2D convolutional layers with the kernel size $7 \times 7$. We use four dense blocks with the size of $(6, 12, 24, 16)$. To preserve more low-level local information, we discard the first max-pooling layer that follows after the first convolution layer. Adam with default parameters and learning rate $\gamma_u = \gamma_l = 0.001$ are used as the optimizer during training. We set Batch Size as only $12$ due to the large memory usage of 3D images. Random crop of $64\times64\times64$, random flip and random transpose are applied as data augmentation.


\begin{figure}[h!]
    \centering
    \includegraphics[width=\columnwidth]{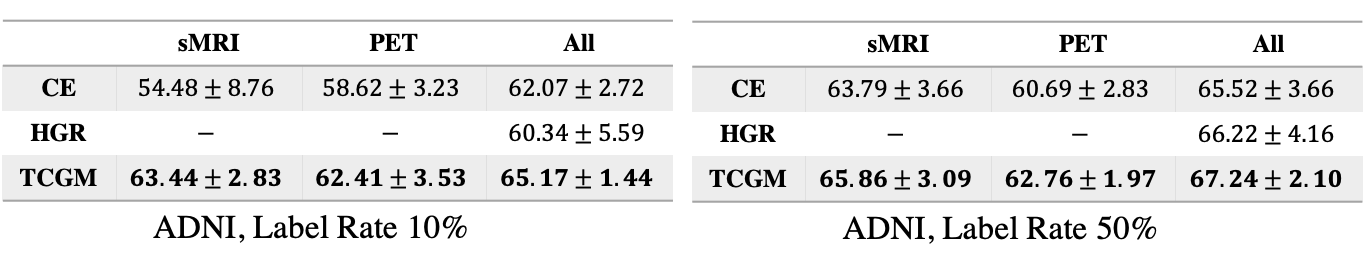}
    \caption{Test accuracies (mean $\pm$ std.) on ADNI dataset}
    \label{fig:adni}
\end{figure}

Figure~\ref{fig:adni} shows the accuracy of classifiers for each modality and the aggregator. Our method \textbf{TCGM} outperforms the baseline methods in all settings especially when the label rate is small, which is desired since it is costly to label data. 

\begin{figure}[h!]
    \centering
    \includegraphics[width=0.8\columnwidth]{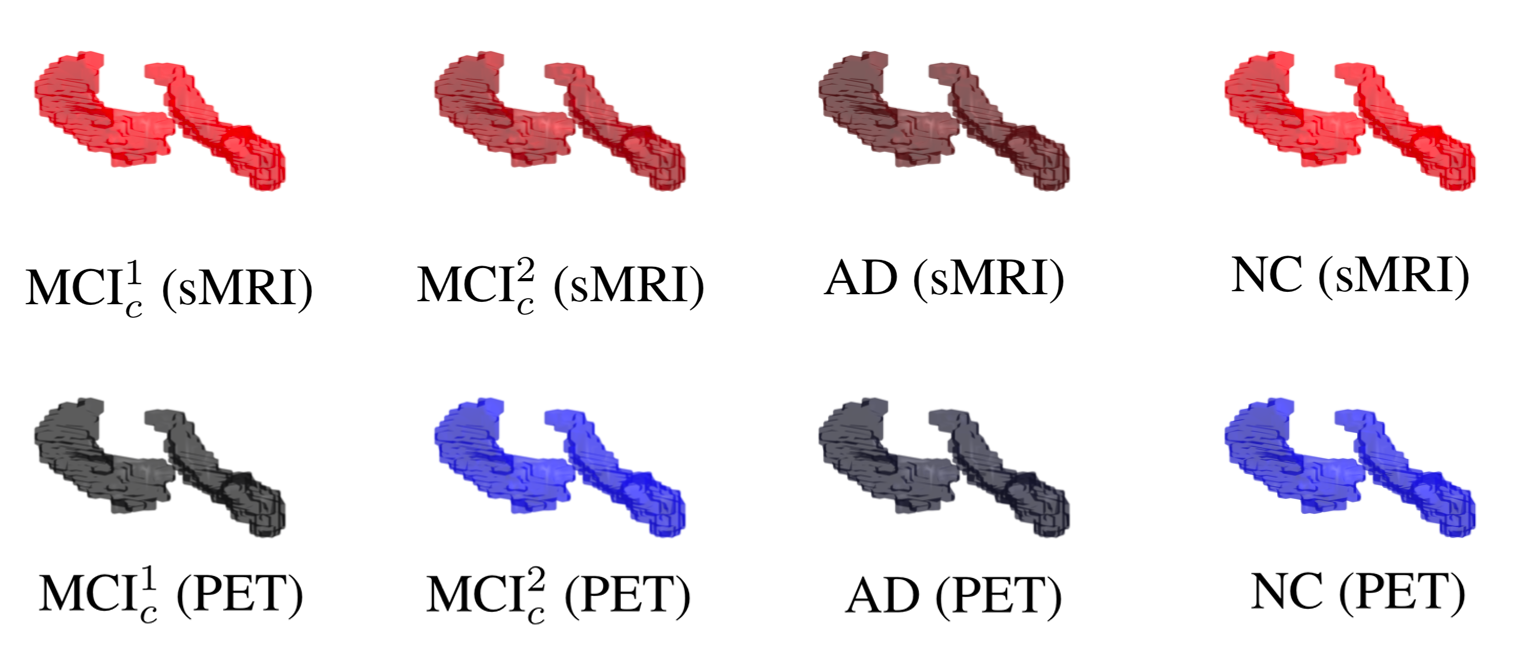}
    \caption{Volume (sMRI, top line) and SUV (PET, bottom line) of MCI$_c^1$, MCI$_c^2$, AD and NC. Darker color implies smaller volume and SUV, i.e., more probability of being AD.}
    \label{fig:ad-visualize}
\end{figure}

To further illustrate the advantage of our model over others in terms of leveraging the knowledge of another modality, we visualize two MCI$_c$s, denoted as MCI$_c^1$ and MCI$_c^2$, which are mistakenly classified as NC by \textbf{CE}'s classifier for sMRI and PET modality, respectively. The volume and standardized uptake value (SUV) (a measurement of the degree of metabolism), whose information are respectively contained by sMRI and PET data, are linearly mapped to the darkness of the red and blue. Darker color implies smaller volume and SUV, i.e., more probability of being AD. As shown in Figure~\ref{fig:ad-visualize}, the volume (SUV) of MCI$_c^1$ (MCI$_c^2$) is similar to NC, hence it is reasonable for \textbf{CE} to mistakenly classify it by only using the information of volume (SUV). In contrast, \textbf{TCGM} for each modality can correctly classify both cases as AD, which shows the better learning of the information intersection (i.e., the ground truth) during training, facilitated by the leverage of knowledge from another modality.

\section{Conclusion}

In this paper, we propose an information-theoretic framework on multi-modal data, Total Correlation Gain Maximization (TCGM), in the scenario of semi-supervised learning. Specifically, we learn to infer the ground truth labels shared by all modalities by maximizing the total correlation gain. Conditioning on a common assumption that all modalities are independent given the ground truth label, it can be theoretically proved our method can learn the Bayesian posterior classifier for each modality and the Bayesian posterior aggregator for all modalities. Extensive experiments on Newsgroup, IEMOCAP, MOSI and ADNI datasets are conducted and achieve promising results, which demonstrates the benefit and utility of our framework.

\section*{Acknowledgement} Yizhou Wang's work is supported by MOST-2018AAA0102004, NSFC-61625201, DFG TRR169/NSFC Major International Collaboration Project "Crossmodal Learning". Yuqing Kong's work is supported by Science and Technology Innovation 2030 –"The New Generation of Artificial Intelligence" Major Project No.2018AAA0100901, China.

\newpage

\bibliographystyle{splncs04}
\bibliography{ref.bib}

\begin{thebibliography}{10}
\providecommand{\url}[1]{\texttt{#1}}
\providecommand{\urlprefix}{URL }
\providecommand{\doi}[1]{https://doi.org/#1}

\bibitem{achille2018emergence}
Achille, A., Soatto, S.: Emergence of invariance and disentanglement in deep
  representations. The Journal of Machine Learning Research  \textbf{19}(1),
  1947--1980 (2018)

\bibitem{ashburner2007fast}
Ashburner, J.: A fast diffeomorphic image registration algorithm. Neuroimage
  \textbf{38}(1),  95--113 (2007)

\bibitem{balcan2005co}
Balcan, M.F., Blum, A., Yang, K.: Co-training and expansion: Towards bridging
  theory and practice. In: Advances in neural information processing systems.
  pp. 89--96 (2005)

\bibitem{belghazi2018mine}
Belghazi, M.I., Baratin, A., Rajeswar, S., Ozair, S., Bengio, Y., Courville,
  A., Hjelm, R.D.: Mine: mutual information neural estimation. arXiv preprint
  arXiv:1801.04062  (2018)

\bibitem{bisson2012co}
Bisson, G., Grimal, C.: Co-clustering of multi-view datasets: a parallelizable
  approach. In: 2012 IEEE 12th International Conference on Data Mining. pp.
  828--833. IEEE (2012)

\bibitem{blum1998combining}
Blum, A., Mitchell, T.: Combining labeled and unlabeled data with co-training.
  In: Proceedings of the eleventh annual conference on Computational learning
  theory. pp. 92--100. ACM (1998)

\bibitem{Busso2008IEMOCAPIE}
Busso, C., Bulut, M., Lee, C.C., Kazemzadeh, A., Provost, E.M., Kim, S., Chang,
  J.N., Lee, S., Narayanan, S.S.: Iemocap: interactive emotional dyadic motion
  capture database. Language Resources and Evaluation  \textbf{42},  335--359
  (2008)

\bibitem{cao2018max}
Cao, P., Xu, Y., Kong, Y., Wang, Y.: Max-mig: an information theoretic approach
  for joint learning from crowds. ICLR2019  (2018)

\bibitem{chandar2016correlational}
Chandar, S., Khapra, M.M., Larochelle, H., Ravindran, B.: Correlational neural
  networks. Neural computation  \textbf{28}(2),  257--285 (2016)

\bibitem{chang2018scalable}
Chang, X., Xiang, T., Hospedales, T.M.: Scalable and effective deep cca via
  soft decorrelation. In: Proceedings of the IEEE Conference on Computer Vision
  and Pattern Recognition. pp. 1488--1497 (2018)

\bibitem{cheng2015semi}
Cheng, Y., Zhao, X., Huang, K., Tan, T.: Semi-supervised learning and feature
  evaluation for rgb-d object recognition. Computer Vision and Image
  Understanding  \textbf{139},  149--160 (2015)

\bibitem{christoudias2006co}
Christoudias, C.M., Saenko, K., Morency, L.P., Darrell, T.: Co-adaptation of
  audio-visual speech and gesture classifiers. In: Proceedings of the 8th
  international conference on Multimodal interfaces. pp. 84--91. ACM (2006)

\bibitem{dasgupta2002pac}
Dasgupta, S., Littman, M.L., McAllester, D.A.: Pac generalization bounds for
  co-training. In: Advances in neural information processing systems. pp.
  375--382 (2002)

\bibitem{hjelm2018learning}
Hjelm, R.D., Fedorov, A., Lavoie-Marchildon, S., Grewal, K., Trischler, A.,
  Bengio, Y.: Learning deep representations by mutual information estimation
  and maximization. arXiv preprint arXiv:1808.06670  (2018)

\bibitem{huang2015euclidean}
Huang, S.L., Suh, C., Zheng, L.: Euclidean information theory of networks. IEEE
  Transactions on Information Theory  \textbf{61}(12),  6795--6814 (2015)

\bibitem{hussain2010icmla}
Hussain, S.F., Grimal, C., Bisson, G.: An improved co-similarity measure for
  document clustering. In: International Conference on Machine Learning and
  Applications (December 2010)

\bibitem{jones2005learning}
Jones, R.: Learning to extract entities from labeled and unlabeled text. Ph.D.
  thesis, Citeseer (2005)

\bibitem{kim2017multi}
Kim, D.H., Lee, M.K., Choi, D.Y., Song, B.C.: Multi-modal emotion recognition
  using semi-supervised learning and multiple neural networks in the wild. In:
  Proceedings of the 19th ACM International Conference on Multimodal
  Interaction. pp. 529--535. ACM (2017)

\bibitem{kong2018water}
Kong, Y., Schoenebeck, G.: Water from two rocks: Maximizing the mutual
  information. In: Proceedings of the 2018 ACM Conference on Economics and
  Computation. pp. 177--194. ACM (2018)

\bibitem{leskes2005value}
Leskes, B.: The value of agreement, a new boosting algorithm. In: International
  Conference on Computational Learning Theory. pp. 95--110. Springer (2005)

\bibitem{levin2003unsupervised}
Levin, A., Viola, P.A., Freund, Y.: Unsupervised improvement of visual
  detectors using co-training. In: ICCV. vol.~1, p.~2 (2003)

\bibitem{lewis1998naive}
Lewis, D.D.: Naive (bayes) at forty: The independence assumption in information
  retrieval. In: European conference on machine learning. pp. 4--15. Springer
  (1998)

\bibitem{Liu2018EfficientLM}
Liu, Z., Shen, Y., Lakshminarasimhan, V.B., Liang, P.P., Zadeh, A., Morency,
  L.P.: Efficient low-rank multimodal fusion with modality-specific factors.
  In: ACL (2018)

\bibitem{vat}
Miyato, T., ichi Maeda, S., Koyama, M., Ishii, S.: Virtual adversarial
  training: a regularization method for supervised and semi-supervised
  learning. In: IEEE transactions on pattern analysis and machine intelligence
  (2018). (2018)

\bibitem{Ng2001OnSC}
Ng, A.Y., Jordan, M.I., Weiss, Y.: On spectral clustering: Analysis and an
  algorithm. In: NIPS (2001)

\bibitem{ngiam2011multimodal}
Ngiam, J., Khosla, A., Kim, M., Nam, J., Lee, H., Ng, A.Y.: Multimodal deep
  learning. In: Proceedings of the 28th international conference on machine
  learning (ICML-11). pp. 689--696 (2011)

\bibitem{nguyen2010estimating}
Nguyen, X., Wainwright, M.J., Jordan, M.I.: Estimating divergence functionals
  and the likelihood ratio by convex risk minimization. IEEE Transactions on
  Information Theory  \textbf{56}(11),  5847--5861 (2010)

\bibitem{sohn2014improved}
Sohn, K., Shang, W., Lee, H.: Improved multimodal deep learning with variation
  of information. In: Advances in Neural Information Processing Systems. pp.
  2141--2149 (2014)

\bibitem{studeny1998multiinformation}
Studen{\`y}, M., Vejnarov{\'a}, J.: The multiinformation function as a tool for
  measuring stochastic dependence. In: Learning in graphical models, pp.
  261--297. Springer (1998)

\bibitem{velivckovic2018deep}
Veli{\v{c}}kovi{\'c}, P., Fedus, W., Hamilton, W.L., Li{\`o}, P., Bengio, Y.,
  Hjelm, R.D.: Deep graph infomax. arXiv preprint arXiv:1809.10341  (2018)

\bibitem{wang2018efficient}
Wang, L., Wu, J., Huang, S.L., Zheng, L., Xu, X., Zhang, L., Huang, J.: An
  efficient approach to informative feature extraction from multimodal data.
  arXiv preprint arXiv:1811.08979  (2018)

\bibitem{Xu2020ATO}
Xu, Y., Zhao, S., Song, J., Stewart, R.J., Ermon, S.: A theory of usable
  information under computational constraints. ArXiv  \textbf{abs/2002.10689}
  (2020)

\bibitem{yang2018semi}
Yang, Y., Zhan, D.C., Sheng, X.R., Jiang, Y.: Semi-supervised multi-modal
  learning with incomplete modalities. In: IJCAI. pp. 2998--3004 (2018)

\bibitem{Zadeh2016MOSIMC}
Zadeh, A., Zellers, R., Pincus, E., Morency, L.P.: Mosi: Multimodal corpus of
  sentiment intensity and subjectivity analysis in online opinion videos. ArXiv
   \textbf{abs/1606.06259} (2016)

\end{thebibliography}

\newpage

\appendix
\section{Proofs}

\begin{proof}[proof of Lemma~\ref{lemma:optimal-reward}]
Note that the expected form of Eq.~\eqref{eq:emp-tc} is the right hand side of Eq.~\eqref{eq:dv}, since $\{x_i^{[M]}\}_i$ are i.i.d from joint distribution $U_{X^{[M]}}$. Therefore, from Lemma~\ref{lemma:dual-tc}, one can immediately get the conclusion. 
\end{proof}

\begin{lemma}
\label{lemma:generating}
Given a joint distribution $p(x^1,...,x^M, y)$, where y is a discrete random variable, we can always find $M$ independent random variables $z^1,...,z^M$ such that $z^i \perp \!\!\! \perp y$ and $x^m = f_m(y,z^m)$, for $m \in [M]$.
\end{lemma}

\begin{proof}
This proof is a generalization of the Proposition in \cite{achille2018emergence}. For $z_1,...,z_M \overset{i.i.d}{\sim}  Uniform(0, 1)$, then from \cite{achille2018emergence} we have $x^m|y = F_{y,m}^{-1}(z_m)$ where $F_{y,m}(t) = \mbox{P}(x^m \leq t | y)$ where $\mbox{P}(x^m \leq t | y)$ is the cumulative distribution function of $p(x^m | y)$. 
\end{proof}

\begin{lemma}
\label{lemma:PTC}
Given assumption~\ref{cond}, then the Marginal-joint ratio (definition~\ref{def:ptc}) $R(x^1,...,x^M)$ has
\begin{align*}
    R(x^1,...,x^M) = \sum_{y \in \mathcal{C}} \frac{ \Pi_{m} p(y | x^m)}{ p(y)^{M-1}}
\end{align*}
Further, the optimal $g$ to make the equality holds in Lemma~\ref{lemma:dual-tc} has 
\begin{align*}
    g(x^1,...,x^M) = 1 + \log{\sum_{y \in \mathcal{C}} \frac{ \Pi_{m} p(y | x^m)}{ p(y)^{M-1}}}.
\end{align*}
\end{lemma}

\begin{proof}

\begin{align*}
    R(x^1,...,x^M) & = \frac{p(x^1,...,x^M)}{\Pi_m p(x^m)} \\
                   & = \frac{\sum_y p(x^1,...,x^M,y)}{\Pi_m p(x^m)} \\
                   & = \frac{\sum_y p(x^1,...,x^M | y)p(y)}{\Pi_m p(x^m)} \\
                   & = \frac{\sum_y \pi_m p(x^1 | y)p(y)}{\Pi_m p(x^m)} \\
                   & = \sum_y  \frac{\pi_m p(y|x^m)}{p(y)^{M-1}}
\end{align*}
One can immediately gets the conclusion for $g$ from Lemma~\ref{lemma:dual-tc}.
\end{proof}

\begin{theorem}[Main theorem] 
Define the expected total correlation gain $TCg(h^1,...,h^M,p)$ as
\begin{equation*}
    TCg(h^1,...,h^M,p) = \mathbb{E}_{x_i^{[M]}\xleftarrow{\text{i.i.d.}} U_{X^{[M]}}} \left(\mathcal{L}_{\text{TC}}(x_i^{[M]}; h^{[M]},\mathbf{p}) \right)
\end{equation*}
Given the conditional independence assumption~\ref{cond} and well-defined prior assumption~\ref{prior-well}, we have that 
\paragraph{Ground-truth $\to$ Maximizer} $(h_{*}^{[M]}, \mathbf{p}^{*})$ is a maximizer of $\max_{\forall m, h^m\in H^m, \mathbf{p} \in \Delta_{ \mathcal{C}}} TCg(h^1,...,h^M,p)$, in other words, $\forall h^{[M]}\in H^{[M]}, \mathbf{p} \in \Delta_{ \mathcal{C}}$,
\begin{align*}
    TCg(h_{\star}^1,...,h_{\star}^M,p^{\star}) \geq TCg(h^1,...,h^M,p^{\star})
\end{align*}
\paragraph{Maximizer $\to$ (Permuted) Ground-truth } If the prior is well defined, then for any maximizer of $TCg$, $(\tilde{h}^{[M]}, \tilde{\mathbf{p}})$, there is a permutation $\tilde{\pi}: \mathcal{C} \to \mathcal{C}$ such that:
\begin{align*}
    \tilde{h}^m(x^m)_c = \mbox{P}(Y = \tilde{\pi}(c) | X^m = x^m), \  \tilde{\mathbf{p}}_c = \mbox{P}(Y = \tilde{\pi}(c))
\end{align*}
\end{theorem}

\begin{proof}
We have
\begin{align}
  TCg(h^1,...,h^M,p) & = \mathbb{E}_{x^{[M]} \gets U_{X^{[M]}}} ( 1 +  \log{\sum_{y \in \mathcal{C}} \frac{ \Pi_{m} h_{\star}^m(x^m) }{ \mathbf{p}^{\star}(y)^{M-1} } } ) \nonumber \\
    & - \mathbb{E}_{x^{[M]} \gets V_{X^{[M]}}}  \sum_{y \in \mathcal{C}} \frac{ \Pi_{m} h_{\star}^m(x^m_i) }{ \mathbf{p}^{\star}(y)^{M-1} }  
\end{align}
\paragraph{Ground-truth $\to$ Maximizer} From definition~\ref{cond}, i.e., 
\begin{equation*}
    h_{\star}^m(x^m)_c = \mbox{P}(Y = c | x^m), \ (p^{\star})_c = \mbox{P}(Y = c)
\end{equation*}
Inspired by Lemma~\ref{lemma:optimal-reward} and~\ref{lemma:dual-tc}, we have that the $TCg(h_{*}^{[M]},p^{\star})$ can achieve the maximum value, which equals to $\tc(X^1,...,x^M)$.

\paragraph{Maximizer $\to$ (Permuted) Ground-truth } For any maximizer $(\tilde{h}^{[M]}, \tilde{\mathbf{p}})$, we have from Lemma~\ref{lemma:optimal-reward} that
\begin{align*}
    \mathcal{R}(\tilde{h}^{[M]}, \tilde{\mathbf{p}}) = 1 + \ptc(x^1,...,x^M),
\end{align*}
which means that the $(\tilde{h}^{[M]}, \tilde{\mathbf{p}})$ satisfies Eq.~\eqref{eq:prior-condition}. With assumption~\ref{prior-well}, we have that there exists a permutation $\tilde{pi}: \mathcal{C} \to \mathcal{C}$ such that
\begin{align*}
    \tilde{h}^m(x^m)_c = \mbox{P}(Y = \tilde{\pi}(c) | X^m = x^m), \  \tilde{\mathbf{p}}_c = \mbox{P}(Y = \tilde{\pi}(c)).
\end{align*}

\end{proof}

\section{Algorithm}

We show the pipeline of TCGM in Alg~\ref{alg: train}.

\begin{algorithm}[h]
\caption{\textbf{TCGM} Optimization}
\label{alg: train}
\begin{algorithmic}
\REQUIRE Unlabeled dataset $\mathcal{D}_u = \{x_i^{[M]}\}_i$, labeled dataset $\mathcal{D}_l = \{(x_i^{[M]}, y_i) \}_i$, $M$ classifiers $\{h^m(\cdot;\Theta^m)\}_{m=1}^M$, epoch number $T$,  learning rate $\gamma_u, \gamma_l$, batch size $N$ and hyperparameter $\mathbf{p}$.

\FOR{epoch $t=1 \to T$}
\FOR{$m=1 \to M$}
\FOR{batch $b = 1 \to \lceil |\mathcal{D}_l|/B \rceil$}
\STATE Randomly sample a batch of samples: \\ $\mathcal{B}_l = \{(x_i^{[M]},y_i)\}_{i=1}^B$ from $\mathcal{D}_l$
\STATE Compute the $\mathcal{L}_{\text{CE}}$ loss: \\ $L \leftarrow \mathcal{L}_{\text{CE}}(\mathcal{B}_l;h^m(\cdot;\Theta^m))$ 
\STATE Update $\Theta^m$: $\Theta^m \leftarrow \Theta^m - \gamma_l \frac{\partial{L}}{\partial{\Theta^m}}$ 
\ENDFOR
\ENDFOR
\FOR{batch $b = 1 \to \lceil (|\mathcal{D}_u|+|\mathcal{D}_l|)/B \rceil$}
\STATE Randomly sample a batch of samples: \\ $\mathcal{B}_{u\cup l} = \{x_i^{[M]}\}_{i=1}^B$ from $\mathcal{D}_u \cup \mathcal{D}_l$
\STATE Compute the $\mathcal{L}_{\text{TC}}$ loss:\\ $L \leftarrow \mathcal{L}_{\textbf{TC}}(\mathcal{B}_{u\cup l};\{h^m(\cdot;\Theta^m)\}_{m=1}^M, \mathbf{p})$ 
\STATE Update $\Theta^{[M]}$: $\forall m, \Theta^m \leftarrow \Theta^m - \gamma_u \frac{\partial{L}}{\partial{\Theta^m}}$ 
\ENDFOR
\ENDFOR
\end{algorithmic}
\end{algorithm}

\paragraph{Time Complexity} The overall loss function of our TCGM method is composed of $\mathcal{L}_{\text{CE}}$ which is $\mathcal{O}(M)$ since it is repeatedly implemented for $M$ classifiers and $\mathcal{L}_{\text{TC}}^{(B)}$, as the addition of two terms, namely the term (a) ($\frac{1}{N}\sum_{i} \log{\sum_{c \in \mathcal{C}} \frac{ \Pi_{m} h^m(x^m_i)_c }{ (p_c)^{M-1} } }$) and the term (b) ($\frac{1}{N!/(N-M)!}\sum_{i_1\neq i_2 \neq \cdots \neq i_M} \sum_{c \in \mathcal{C}} \frac{ \Pi_{m} h^m(x^m_{i_m})_c }{ (p_c)^{M-1} }$) in Eq.~\eqref{eq:empi}. The term (a) with $N$ samples generated from joint distribution $p(x^1,…,x^M)$ for each modality; hence, the optimization of the term (a) is $\mathcal{O}(M)$. For term (b) with $N!/(N-M)!$ samples generated from marginal distribution $\Pi_{m} p(x^m)$, it is the sum of $N$ terms by grouping terms with $h^m(x_i^m)$ for each $i \in [N]$ with $h^m(x_1^m),…,h^m(x_N^m) for m \in [M]$ calculated ahead (which is $\mathcal{O}(M)$), hence is linear scale with respect to $M$. Therefore, the time complexity for our loss function is $\mathcal{O}(M)$.

\section{Extended experiments results}
\label{ref:extent-results}
We show the complete result of single modality classifiers on Emotion Recognition task in Figure~6.

\begin{figure*}[h!]
    \centering
    \subfigure[Happy (text)]{\includegraphics[width=0.3\textwidth]{img/ie_happy_text.pdf}}
    \subfigure[Happy (visual)]{\includegraphics[width=0.3\textwidth]{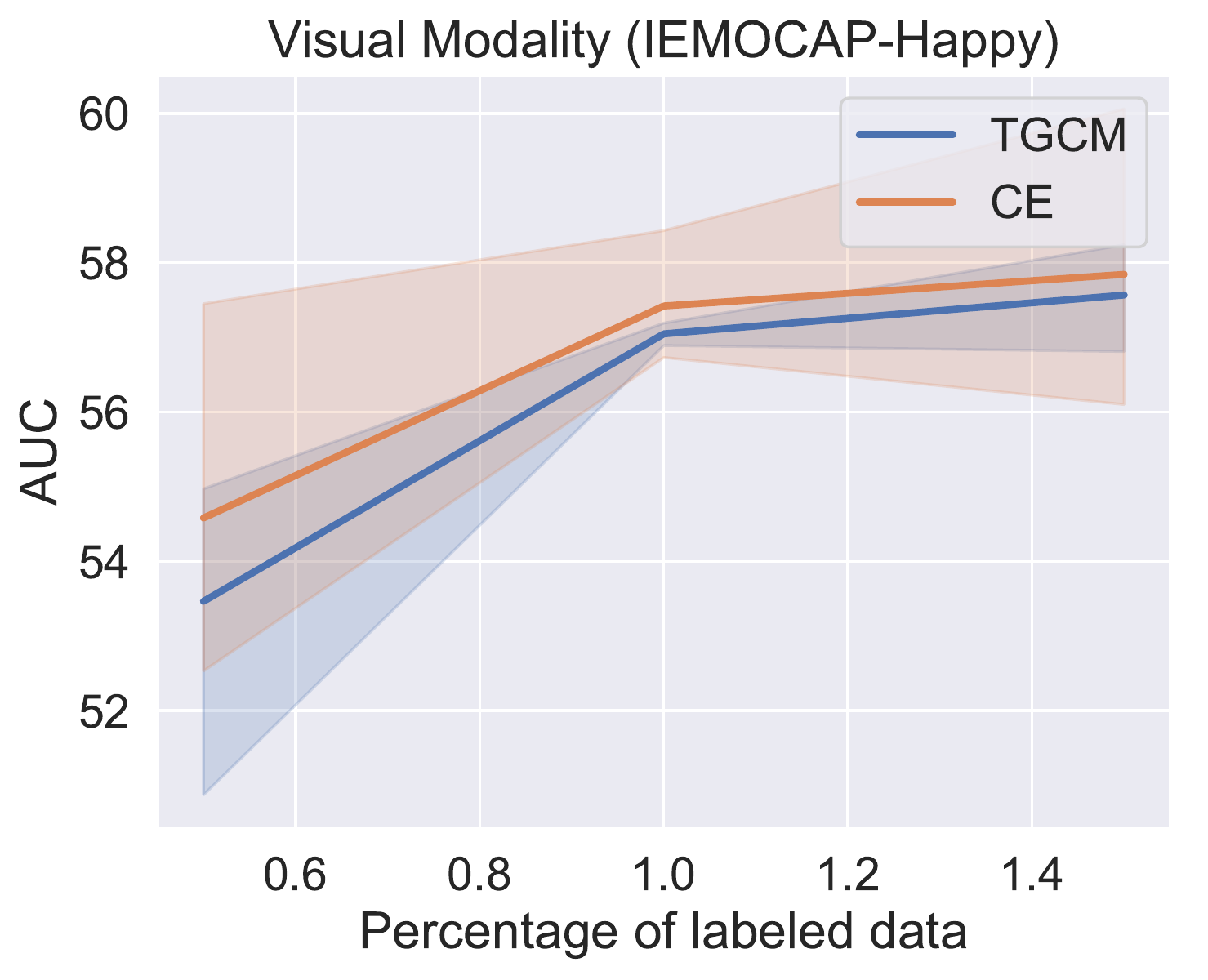}}
    \subfigure[Happy (audio)]{\includegraphics[width=0.3\textwidth]{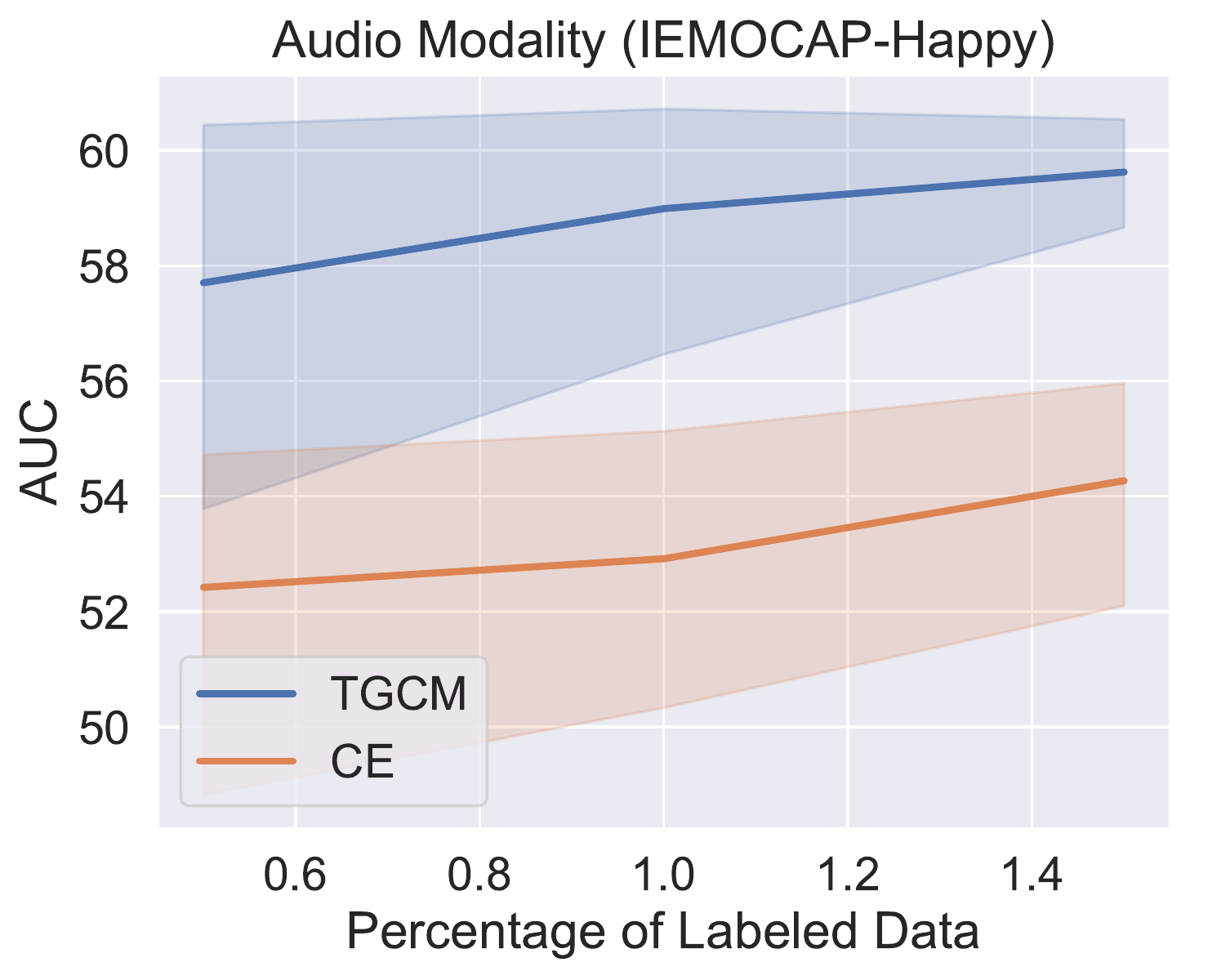}}
    \subfigure[Angry (text)]{\includegraphics[width=0.3\textwidth]{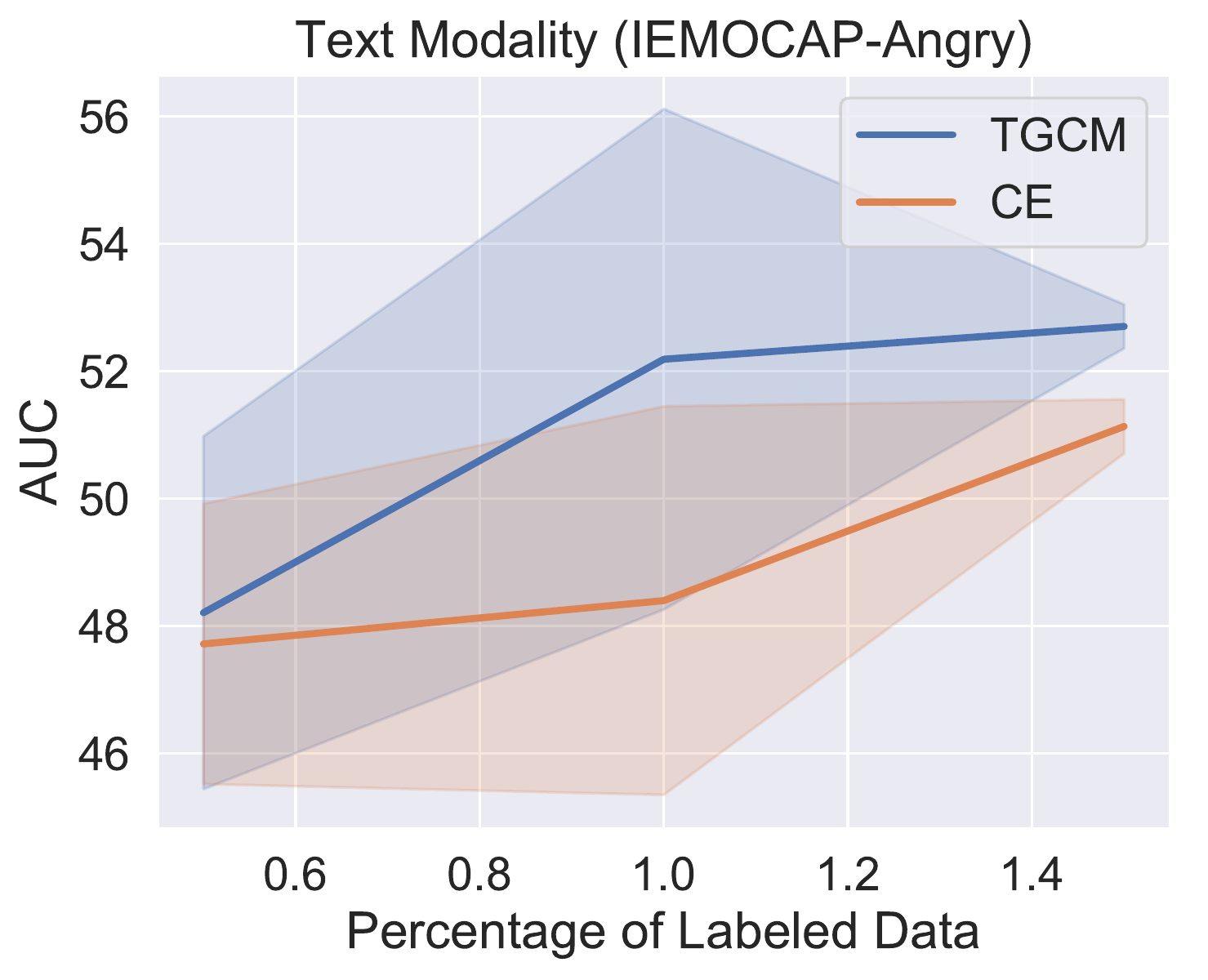}}
    \subfigure[Angry (audio)]{\includegraphics[width=0.3\textwidth]{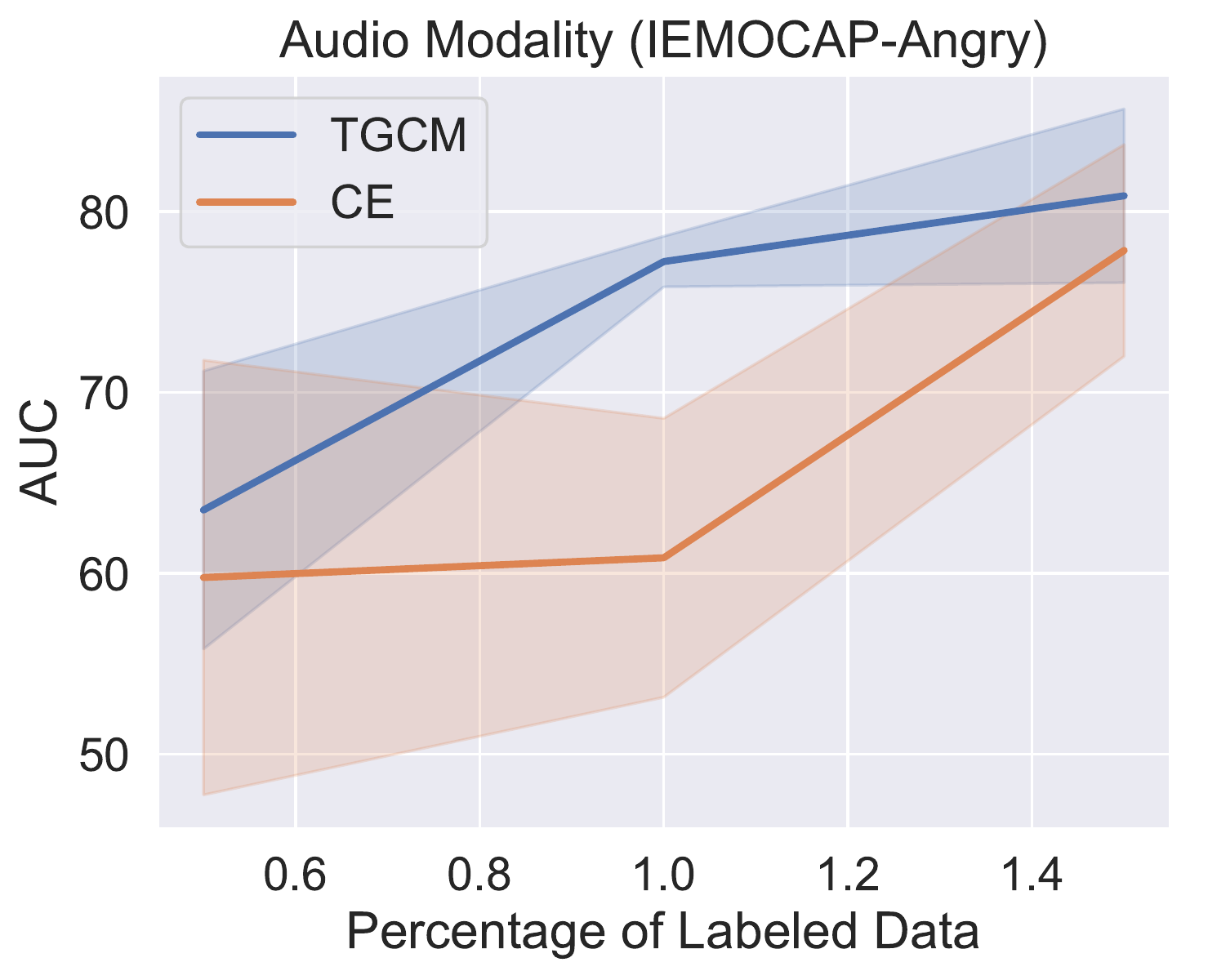}}
    \subfigure[Neutral (text)]{\includegraphics[width=0.3\textwidth]{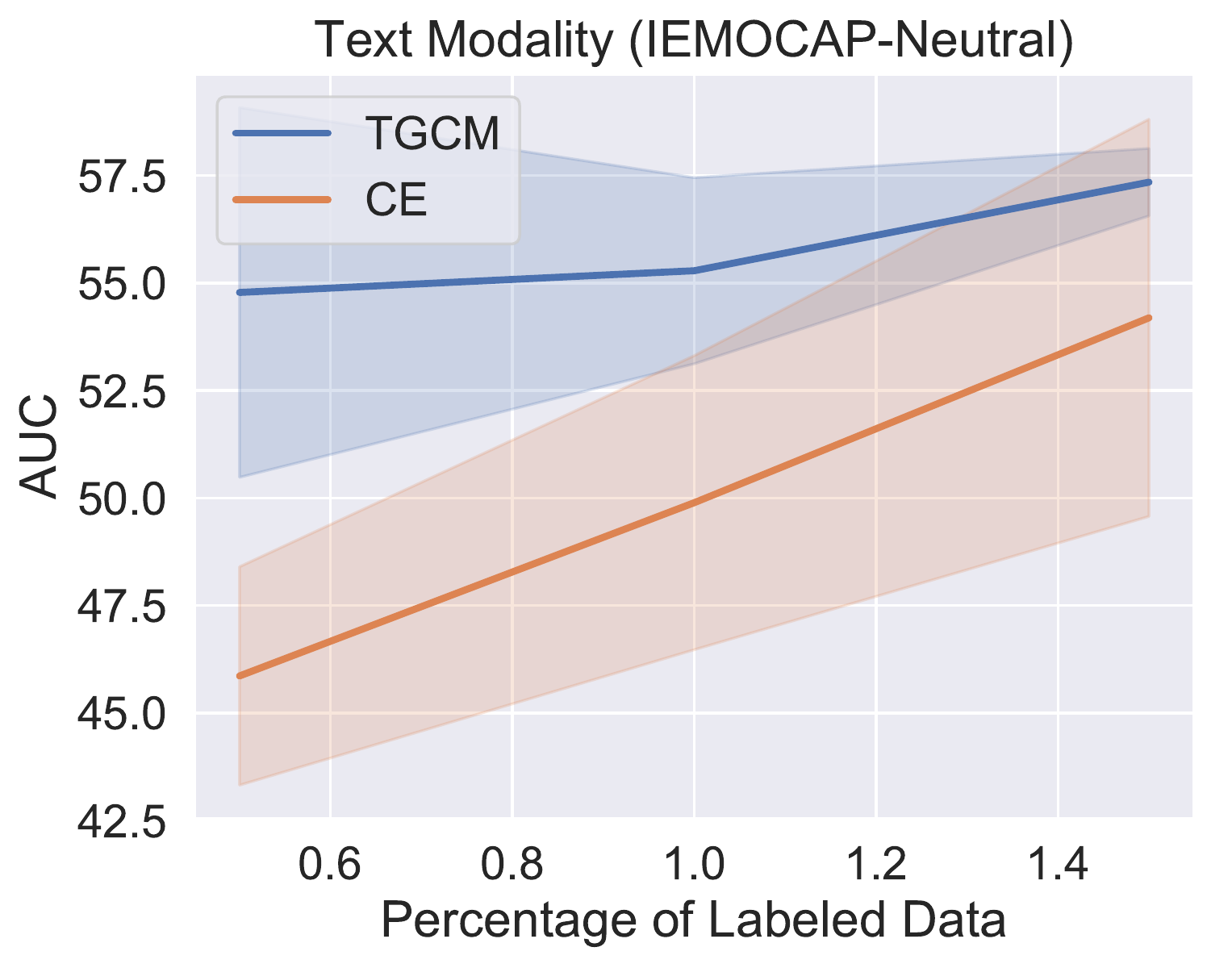}}
    \subfigure[Neutral (video)]{\includegraphics[width=0.3\textwidth]{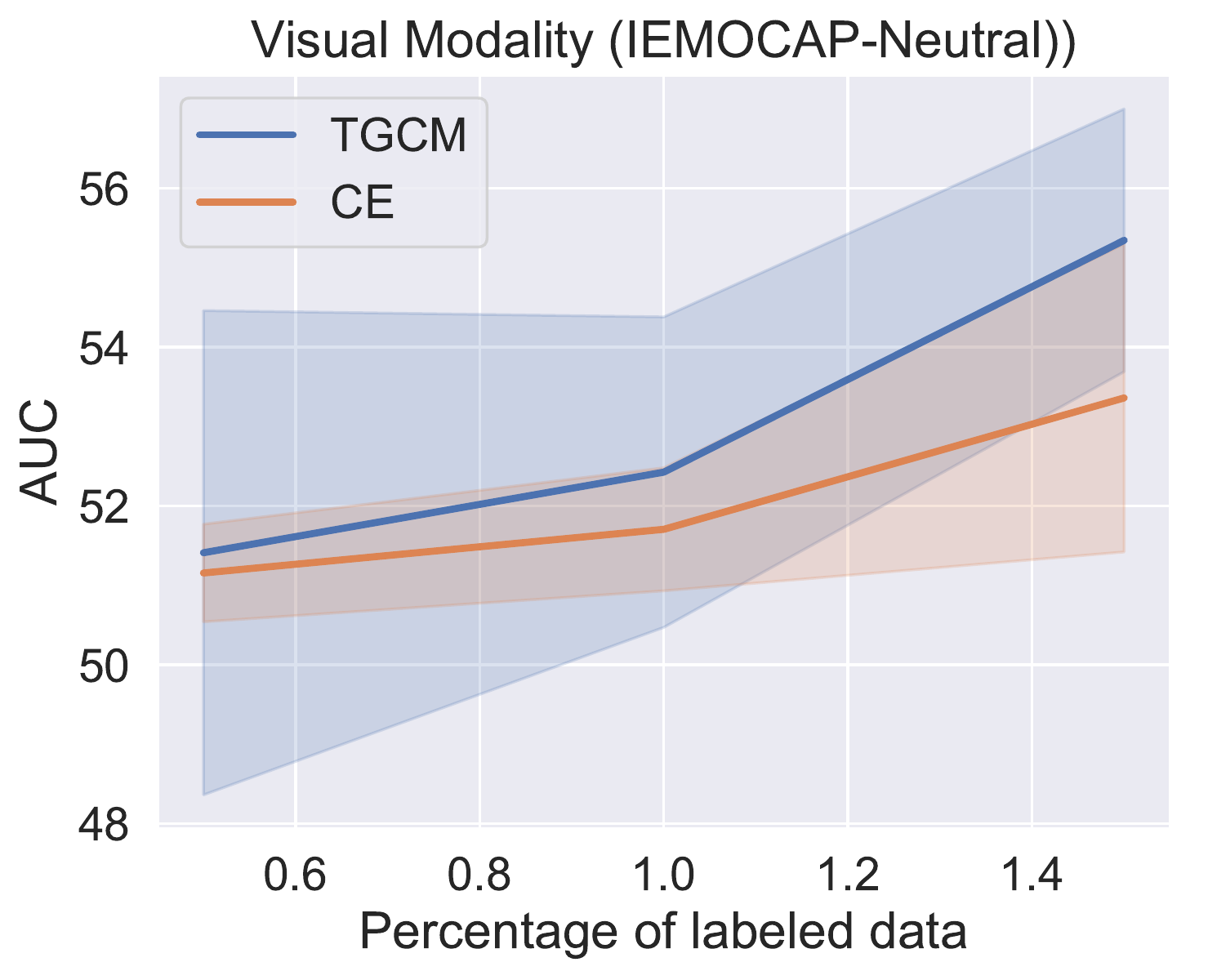}}
    \subfigure[Neutral (audio)]{\includegraphics[width=0.3\textwidth]{img/ie_neutral_audio.pdf}}
    \subfigure[MOSI (text)]{\includegraphics[width=0.3\textwidth]{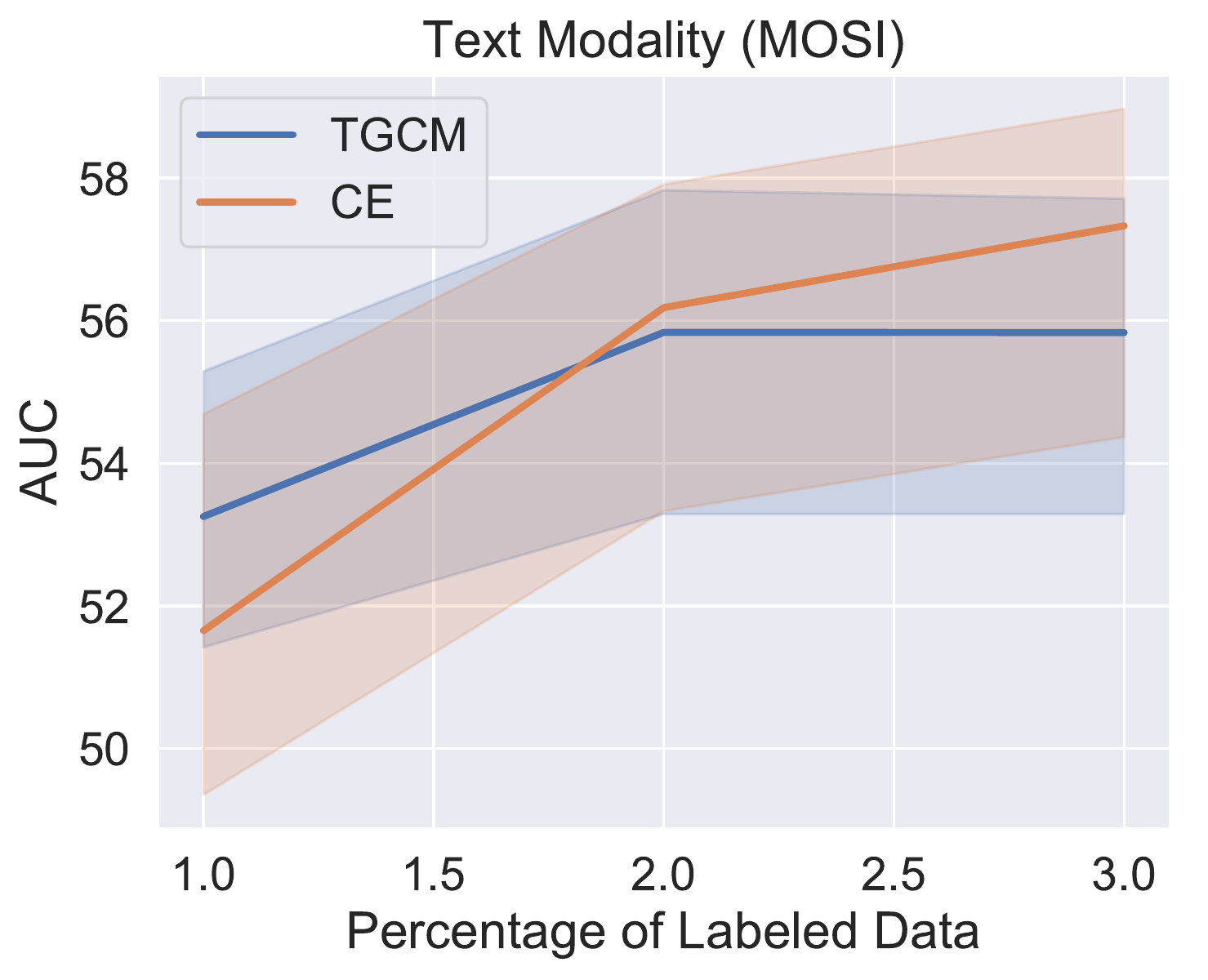}}
    \caption{AUC of single modality classifiers by CE and TCGM.}
\end{figure*}

\end{document}